\def\year{2020}\relax
\documentclass[letterpaper]{article} 
\usepackage{aaai20}  
\usepackage{times}  
\usepackage{helvet} 
\usepackage{courier}  
\usepackage[hyphens]{url}  
\usepackage{graphicx} 
\usepackage{multicol}
\usepackage{graphicx}  
\usepackage[utf8]{inputenc} 
\usepackage[T1]{fontenc}    
\usepackage{hyperref}       
\usepackage{url}            
\usepackage{booktabs}       
\usepackage{amsfonts}       
\usepackage{nicefrac}       
\usepackage{microtype}      

\usepackage{stmaryrd}

\usepackage{graphicx}
\usepackage{booktabs,stix} 
\usepackage{mathtools}

\usepackage{makeidx}  
\usepackage{pgfgantt}
\usepackage{xcolor}
\usepackage{epstopdf}
\usepackage{microtype}
\usepackage{pstricks,pst-plot,pst-text,pst-tree,pst-eps,pst-fill,pst-node,pst-math}
\usepackage{tikz}
\usepackage{pgfplots}	

\usepackage{pgfplots}
\pgfplotsset{compat=newest}
\pgfplotsset{/pgfplots/error bars/error bar style={very thick}}
\pgfplotsset{
  every axis plot/.append style={very thick, black},
}

\usepackage{booktabs} 
\usepackage{multirow} 
\usepackage{multicol} 
\usepackage{amsmath,amsthm,dsfont,tikz,ifthen,subfigure,booktabs,slashbox
}
\usepackage[mathscr]{euscript}
\usepackage{algorithm,algorithmic}
\usepackage{natbib}
\usepackage{wrapfig}
\usepackage{comment}
\newcommand{\Prob}{\vec{p}}

\renewcommand{\Pr}{\mathbf{p}}
\renewcommand{\vec}[1]{\boldsymbol{#1}}
\newcommand{\given}{\, | \,}

\renewcommand{\vec}[1]{\boldsymbol{#1}}

\newcommand*{\defeq}{\mathrel{\vcenter{\baselineskip0.5ex \lineskiplimit0pt
			\hbox{\footnotesize.}\hbox{\footnotesize.}}}%
	=}
\newcommand{\defi}{\defeq}
\usepackage{figsize}

\usepackage{etoolbox}
\AtBeginEnvironment{figure}{\setcounter{subfigure}{0}}

\newtheorem{proposition}{Proposition}
\newtheorem{lemma}{Lemma}
\newtheorem{remark}{Remark}
\newtheorem{corollary}{Corollary}

\title{Reliable Multilabel Classification: Prediction with Partial Abstention}
\author{Vu-Linh Nguyen, Eyke H\"ullermeier\\ 
Heinz Nixdorf Institute and Department of Computer Science\\
Paderborn University, Germany\\ 
vu.linh.nguyen@uni-paderborn.de, eyke@upb.de 
}
\begin{document}

\maketitle

\begin{abstract}
In contrast to conventional (single-label) classification, the setting of \emph{multilabel classification} (MLC) allows an instance to belong to several classes simultaneously. Thus, instead of selecting a single class label, predictions take the form of a subset of all labels. In this paper, we study an extension of the setting of MLC, in which the learner is allowed to partially abstain from a prediction, that is, to deliver predictions on some but not necessarily all class labels. We propose a formalization of MLC with abstention in terms of a generalized loss minimization  problem and present first results for the case of the Hamming loss, rank loss, and F-measure, both theoretical and experimental. 
\end{abstract}

\section{Introduction}
\label{intro}

In statistics and machine learning, classification with abstention, also known as classification with a reject option, is an extension of the standard setting of classification, in which the learner is allowed to refuse a prediction for a given query instance; research on this setting dates back to early work by \citet{chow_oo70} and \citet{hellman:1970} and remains to be an important topic till today, most notably for binary classification \citep{bartlett:2008,cort_lw16,franc:2019,grandvalet:2008}. For the learner, the main reason to abstain is a lack of certainty about the corresponding outcome---refusing or at least deferring a decision might then be better than taking a high risk of a wrong decision. 

Nowadays, there are many machine learning problems in which complex, structured predictions are sought (instead of scalar values, like in classification and regression). For such problems, the idea of abstaining from a prediction can be generalized toward \emph{partial abstention}: Instead of predicting the entire structure, the learner predicts only parts of it, namely those for which it is certain enough. This idea has already been realized, e.g., for the case where predictions are rankings \citep{cheng:2010,cheng:2012}. 

Another important example is \emph{multilabel classification} (MLC), in which an outcome associated with an instance is a labeling in the form of a subset of an underlying reference set of class labels \citep{dembczynski:2012,tsoumakas:2009,zhang:2014}. 
In this paper, we study an extension of the setting of MLC, in which the learner is allowed to partially abstain from a prediction, that is, to deliver predictions on some but not necessarily all class labels (or, more generally, to refuse committing to a single complete prediction). Although MLC has been studied extensively in the machine learning literature in the recent past, there is surprisingly little work on MLC with abstention so far---a notable exception is \citep{pillai:2013}, to which we will return in the Section \ref{sec:rlw}.   

Prediction with abstention is typically realized as a two-stage approach. First, the learner delivers a prediction that provides information about its uncertainty. Then, taking this uncertainty into account, a decision is made about whether or not to predict, or on which parts. In binary classification, for example, a typical approach is to produce probabilistic predictions and to abstain whenever the probability is close to $\nicefrac{1}{2}$. We adopt a similar approach, in which we rely on probabilistic MLC, i.e., probabilistic predictions of labelings.

In the next section, we briefly recall the setting of multilabel classification. The generalization toward MLC with (partial) abstention is then introduced and formalized in Section \ref{sec:MLCA}. 
Instantiations of the setting of MLC with abstention for the specific cases of the Hamming loss, rank loss, and F-measure are studied in Sections \ref{sec:hamming}--\ref{sec:Fmeasure}, respectively. 
Related work is discussed in Section \ref{sec:rlw}. Finally, experimental results are presented in Section \ref{sec:exp}, prior to concluding the paper in Section \ref{sec:con}. All formal results in this paper (propositions, remarks, corollaries) are stated without proofs, which are deferred to the supplementary material. 

\section{Multilabel Classification}
\label{sec:mlc}

In this section, we describe the MLC problem in more detail and formalize it within a probabilistic setting. Along the way, we introduce the notation used throughout the paper.

Let $\mathcal{X}$ denote an instance space, and let $\mathcal{L}= \{\lambda_1, \ldots, \lambda_m\}$ be a finite set of class labels.
We assume that an instance $\vec{x} \in \mathcal{X}$ is (probabilistically) associated with a subset of labels $\Lambda = \Lambda(\vec{x}) \in 2^\mathcal{L}$; this subset is often called the set of relevant labels, while the complement
$\mathcal{L} \setminus \Lambda$ is considered as irrelevant for $\vec{x}$. 
We identify a set $\Lambda$ of relevant labels with a binary vector $\vec{y} = (y_1, \ldots, y_m)$, where $y_i = \llbracket \lambda_i \in \Lambda \rrbracket$.\footnote{$\llbracket \cdot \rrbracket$ is the indicator function, i.e., $\llbracket A \rrbracket = 1$ if the predicate $A$ is true and $=0$ otherwise.} By $\mathcal{Y} = \{0,1\}^m$ we denote the set of possible labelings. 

We assume observations to be realizations of0 random variables generated independently and identically (i.i.d.) according to a probability distribution  $\Prob$ on $\mathcal{X} \times \mathcal{Y}$, i.e., an observation $\vec{y}=(y_1,\ldots, y_m)$ is the realization of a corresponding random vector $\mathbf{Y} = (Y_1, \ldots, Y_m)$. We denote by $\Prob(\mathbf{Y} \given \vec{x})$ the conditional distribution of $\mathbf{Y}$ given $\mathbf{X}=\vec{x}$, and by $\Prob_i(Y_i \given \vec{x})$ the corresponding marginal distribution of~$Y_i$:
\begin{equation}\label{eq:marginal}
\Prob_i( b \given \vec{x}) =  \sum_{\vec{y}\in\mathcal{Y}: y_i = b} \Prob(\vec{y} \given \vec{x}) \, .
\end{equation}
A multilabel classifier $\mathbf{h}$ is a mapping $\mathcal{X} \longrightarrow \mathcal{Y}$  that assigns a (predicted) label subset to each instance $\vec{x}\in \mathcal{X}$. Thus, the output of a classifier $\mathbf{h}$ is a vector 
\begin{equation*}
\hat{\vec{y}} = \mathbf{h}(\vec{x}) = (h_1(\vec{x}), \ldots , h_m(\vec{x})) \,  .
\end{equation*}
Given training data in the form of a finite set of observations
$(\vec{x},\vec{y}) \in \mathcal{X} \times \mathcal{Y}$, drawn independently from $\Pr(\mathbf{X},\mathbf{Y})$, the goal in MLC is to learn a classifier $\mathbf{h}: \, \mathcal{X} \longrightarrow \mathcal{Y}$ that generalizes well beyond these observations in the sense of minimizing the expected risk with respect to a specific loss function. 

In the literature, various MLC loss functions have been proposed, including the Hamming loss, the subset 0/1 loss, the F-measure, the Jaccard measure, and the rank loss. The Hamming loss is given by
\begin{equation}
\label{eq:hamming}
\ell_H(\vec{y}, \hat{\vec{y}}) = 
\sum_{i=1}^m  \, \llbracket y_i \neq  \hat{y}_i \rrbracket \enspace , 
\end{equation}
and the subset 0/1 loss by $\ell_S(\vec{y}, \hat{\vec{y}}) = \llbracket \vec{y} \neq  \hat{\vec{y}} \rrbracket$. Thus, both losses generalize the standard 0/1 loss commonly used in classification, but in a very different way. Hamming and subset 0/1 are prototypical examples of what is called a (label-wise) \emph{decomposable} and \emph{non-decomposable} loss, respectively \citep{dembczynski:2012}. A decomposable loss can be expressed in the form 
\begin{equation}\label{eq:decoml}
\ell(\vec{y} , \hat{\vec{y}} ) = \sum_{i=1}^m \ell_i(y_i , \hat{y_i} )
\end{equation}
with suitable binary loss functions $\ell_i:\, \{0,1\}^2 \longrightarrow \mathbb{R}$, whereas a non-decomposable loss does not permit such a representation. It can be shown that, to produce optimal predictions $\hat{\vec{y}} = \mathbf{h}(\vec{x})$ minimizing expected loss, knowledge about the marginals $\Prob_i(Y_i \given \vec{x})$ is enough in the case of a decomposable loss, but not in the case of a non-decomposable loss \citep{dembczynski:2012}. Instead, if a loss is non-decomposable, high-order probabilities are needed, and in the extreme case even the entire distribution $\Prob(\mathbf{Y} \given \vec{x})$ (like in the case of the subset 0/1 loss). On an algorithmic level, this means that MLC with a decomposable loss can be tackled by what is commonly called binary relevance (BR) learning (i.e., learning one binary classifier for each label individually), whereas non-decomposable losses call for more sophisticated learning methods that are able to take label-dependencies into~account. 

\section{MLC with Abstention}
\label{sec:MLCA}

In our generalized setting of MLC with abstention, which is introduced in this section, the classifier is allowed to produce \emph{partial predictions} 
\begin{equation}\label{eq:papr}
\hat{\vec{y}} = \mathbf{h}(\vec{x}) \in \mathcal{Y}_{pa} \defi \{ 0, \bot , 1 \}^m \, ,
\end{equation}
where $\hat{y}_i = \bot$ indicates an abstention on the label $\lambda_i$; we denote by $A(\hat{\vec{y}}) \subseteq [m] \defi \{1,\ldots , m\}$ and $D(\hat{\vec{y}}) \defi [m] \setminus A(\hat{\vec{y}})$ the set of indices $i$ for which $\hat{y}_i = \bot$ and $\hat{y}_i \in \{0,1\}$, respectively, that is, the indices on which the learner abstains and decides to predict. 

\subsection{Risk Minimization}
\label{sec:MLCRM}

To evaluate a reliable multilabel classifier, a generalized loss function 
\begin{equation}\label{eq:glf}
L:\, \mathcal{Y} \times \mathcal{Y}_{pa} \longrightarrow \mathbb{R}_+
\end{equation}
is needed, which compares a partial prediction $\hat{\vec{y}}$ with a ground-truth labeling $\vec{y}$. 
Given such a loss, and assuming a probabilistic prediction for a query instance $\vec{x}$, i.e., a probability $\Prob(\cdot \given \vec{x})$ on the set of labelings (or at least an estimation thereof), the problem of risk minimization comes down to finding 
\begin{align}\label{eq:glm}
\hat{\vec{y}} &\in \operatorname*{argmin}_{\hat{\vec{y}} \in \mathcal{Y}_{pa}} \mathbf{E} \big( L(\vec{y} , \hat{\vec{y}}) \big)  \\
&= \operatorname*{argmin}_{\hat{\vec{y}} \in \mathcal{Y}_{pa}}\sum_{\vec{y} \in \mathcal{Y}} L(\vec{y} , \hat{\vec{y}}) \cdot \Prob(\vec{y} \given \vec{x}) \, . \nonumber
\end{align}
The concrete form of this optimization problem as well as its difficulty depend on several choices, including the underlying MLC loss function $\ell$ and its extension $L$.

\subsection{Generalized Loss Functions}\label{sec:losspr}

On the basis of a standard MLC loss $\ell$, a generalized loss function (\ref{eq:glf}) can be derived in different ways, also depending on how to penalize the abstention. Further below, we propose a generalization based on an additive penalty. Before doing so, we discuss some general properties that might be of interest for generalized losses. 
  

As a first property, we should expect a generalized loss $L$ to reduce to its conventional version $\ell$ in the case of no abstention. In other words, 
\begin{align*}
L(\vec{y},\hat{\vec{y}}) = \ell(\vec{y}, \hat{\vec{y}} ) \, ,
\end{align*} 
whenever $\hat{\vec{y}}$ is a precise prediction $\hat{\vec{y}} \in \mathcal{Y}$. Needless to say, this is a property that every generalized loss should obey.

\paragraph{Monotonicity.}
Another reasonable property is \emph{monotonicity}: The loss should only increase (or at least not decrease) when 
(i) turning a correct prediction on a label $\lambda_i$ into an abstention or an incorrect prediction, 
(ii) or turning an abstention into an incorrect prediction. 
This reflects the following chain of preferences: a correct prediction is better than an abstention, which in turn is better than an incorrect prediction. 
More formally, for a ground-truth labeling $\vec{y}$ and a partial prediction $\hat{\vec{y}}_1$, let $C_1 , A_1 \subseteq \mathcal{L}$ denote the subset of labels on which the prediction is correct and on which the learner abstains, respectively, and define $C_2 , A_2 \subseteq \mathcal{L}$ analogously for a prediction $\hat{\vec{y}}_2$. Then
\begin{align}\label{eq:mono1}
&(C_2 \subseteq C_1) \; \wedge \;  \Big( (C_2  \cup A_2)   \subseteq  (C_1 \cup A_1) \Big) \\
 & \Rightarrow L(\vec{y}, \hat{\vec{y}}_1) \leq L(\vec{y}, \hat{\vec{y}}_2)  \, . \nonumber
\end{align}

\paragraph{Uncertainty-alignment.}
Intuitively, when producing a partial prediction, an optimal prediction rule is supposed to abstain on the most uncertain labels. More formally, consider a generalized loss function $L$ and a prediction $\hat{\vec{y}}$ which, for a query $\vec{x} \in \mathcal{X}$, is a risk-minimizer (\ref{eq:glm}). Moreover, denoting by $p_i = \Prob_i( 1 \given \vec{x})$ the (marginal) probability that label $\lambda_i$ is relevant for $\vec{x}$, it is natural to quantify the degree of uncertainty on this label in terms~of 
\begin{equation}\label{eq:unc}
u_i = 1-2|p_i - \nicefrac{1}{2} | = 2 \min(p_i,1-p_i) \, ,
\end{equation}
or any other  function symmetric around $\nicefrac{1}{2}$. We say that $\hat{\vec{y}}$ is \emph{uncertainty-aligned}~if
\begin{equation*}
\forall \, y_i \in A(\hat{\vec{y}}) , y_j \in D(\hat{\vec{y}}) : \, 
u_i \geq u_j \, .
\end{equation*}
Thus, a prediction is uncertainty-aligned if the following holds: Whenever the learner decides to abstain on label $\lambda_i$ and to predict on label $\lambda_j$, the uncertainty on  $\lambda_j$ cannot exceed the uncertainty on $\lambda_i$. We then call a loss function $L$ uncertainty-aligned if it guarantees the existence of an uncertainty-aligned risk-minimizer, regardless of the probability $\Prob = \Prob( \cdot \given \vec{x})$. 

\subsubsection{Additive Penalty for Abstention}

Consider the case of a partial prediction $\hat{\vec{y}}$ and denote by $\hat{\vec{y}}_D$ and $\hat{\vec{y}}_A$ the projections of $\hat{\vec{y}}$ to the entries in $D(\hat{\vec{y}})$ and $A(\hat{\vec{y}})$, respectively. As a natural extension of the original loss $\ell$, we propose a generalized loss of the form
\begin{equation}\label{eq:glossa}
L(\vec{y}, \hat{\vec{y}}) =  \ell(\vec{y}_D , \hat{\vec{y}}_D) + f(A(\hat{\vec{y}})) \, ,
\end{equation}
with $\ell(\vec{y}_D , \hat{\vec{y}}_D)$ the original loss on that part on which the learner predicts and $f(A(\hat{\vec{y}}))$ a penalty for abstaining on $A(\hat{\vec{y}})$. The latter can be seen as a measure of the loss of usefulness of the prediction $\hat{\vec{y}}$ due to its partiality, i.e., due to having no predictions on $A(\hat{\vec{y}})$. 

An important instantiation of (\ref{eq:glossa}) is the case where the penalty is a counting measure, i.e., where $f$ only depends on the number of abstentions:
\begin{equation}\label{eq:glosscount}
L(\vec{y}, \hat{\vec{y}}) =  \ell(\vec{y}_D , \hat{\vec{y}}_D) + f\big(|A(\hat{\vec{y}})|\big) \, .
\end{equation}
A special case of \eqref{eq:glosscount} is to penalize each abstention $\hat{y}_i = \bot$ with the same constant $c \in [0,1]$, which yields
\begin{equation}\label{eq:gloss1}
L(\vec{y}, \hat{\vec{y}}) = \ell(\vec{y}_D , \hat{\vec{y}}_D) + |A(\hat{\vec{y}})| \cdot c \, .
\end{equation}
Of course, instead of a linear function $f$, more general penalty functions are conceivable. For example, a practically relevant penalty is a concave function of the number of abstentions: Each additional abstention causes a cost, so $f$ is monotone increasing in $|A(\hat{\vec{y}})|$, but the marginal cost of abstention is~decreasing. 

\begin{proposition}\label{pro:decloss}
Let the loss $\ell$ be decomposable in the sense of \eqref{eq:decoml}, and let $\hat{\vec{y}}$ be a risk-minimizing prediction (for a query instance $\vec{x}$). The minimization of the expected loss \eqref{eq:glosscount} is then accomplished~by 
\begin{equation}\label{eq:hlmx}
\hat{\vec{y}} = \operatorname*{argmin}_{1 \leq d \leq m} 
\mathbf{E} \left(\ell(\vec{y},\hat{\vec{y}}_d) \right) + f(m-d) \, ,
\end{equation}
where the prediction $\hat{\vec{y}}_d$ is specified by the index set 
\begin{equation}\label{eq:hamslx}
D_d(\hat{\vec{y}}_d) \defi  \{\pi(1), \ldots , \pi(d)\} \, ,
\end{equation}
and the permutation $\pi$ sorts the labels in increasing order of the label-wise expected losses 
\begin{align*}
s_i = \min_{\hat{y}_i \in \{0,1\}}\mathbf{E}(\ell_i(y_i, \hat{y}_i)) \, , 
\end{align*}
i.e., $s_{\pi(1)} \leq \cdots \leq s_{\pi(m)}$.
\end{proposition}

As shown by the previous proposition, a risk-minimizing prediction for a decomposable loss can easily be found in time $O(m \log(m))$, simply by sorting the labels according to their contribution to the expected loss, and then finding the optimal size $d$ of the prediction according to~\eqref{eq:hlmx}.

\section{The Case of Hamming Loss}
\label{sec:hamming}

This section presents first results for the case of the Hamming loss function \eqref{eq:hamming}. In particular, we analyze extensions of the Hamming loss according to \eqref{eq:glosscount} and address the corresponding problem of risk minimization. 

Given a query instance $\vec{x}$, assume conditional probabilities 
$p_i = \Prob(y_i = 1 \given \vec{x})$ are given or made available by an MLC predictor $\mathbf{h}$. In the case of Hamming, the expected loss of a prediction $\hat{\vec{y}}$ is then given by
\begin{align*} 
\mathbf{E} \left(\ell_H(\vec{y}, \hat{\vec{y}}) \right)  = 
  \sum_{i : \hat{y}_i = 1} 1-p_i  + \sum_{i : \hat{y}_i = 0} p_i  
\end{align*}
and minimized by $\hat{\vec{y}}$ such that $\hat{y}_i = 0$ if $p_i \leq \nicefrac{1}{2}$ and $\hat{y}_i = 1$ otherwise.

In the setting of abstention, we call a prediction $\hat{\vec{y}}$ a d-prediction if $|D(\hat{\vec{y}})| = d$. Let $\pi$ be a permutation of $[m]$ that sorts labels according to the uncertainty degrees (\ref{eq:unc}), i.e., such that $u_{\pi(1)} \leq u_{\pi(2)} \leq \cdots \leq u_{\pi(m)}$. As a consequence of Proposition \ref{pro:decloss}, we obtain the following result.   
 
\begin{corollary}\label{pro:hamopt}
In the case of Hamming loss, let $\hat{\vec{y}}$ be a risk-minimizing prediction (for a query instance $\vec{x}$). The minimization of the expected loss \eqref{eq:glosscount} is then accomplished by 
\begin{equation}\label{eq:hlm}
\hat{\vec{y}} = \operatorname*{argmin}_{1 \leq d \leq m} 
\mathbf{E} \left(\ell_H(\vec{y}, \hat{\vec{y}}_d) \right) + f(m-d) \, ,
\end{equation}
where the prediction $\hat{\vec{y}}_d$ is specified by the index set 
\begin{equation}\label{eq:hamsl}
D_d(\hat{\vec{y}}_d) =  \{\pi(1), \ldots , \pi(d)\} \, .
\end{equation}
\end{corollary}

\begin{corollary}\label{pro:Hamming1}
The extension \eqref{eq:glosscount} of the Hamming loss is uncertainty-aligned. In the case of the extension \eqref{eq:gloss1} of the Hamming loss, the optimal prediction is given by \eqref{eq:hamsl} with 
\begin{align*}
d = | \{ i \in [m] \given \min\left(p_i , 1- p_i \right) \leq c\} | \,.
\end{align*} 
\end{corollary}

\begin{remark}\label{prop1}
The extension \eqref{eq:glosscount} of the Hamming loss is monotonic, provided $f$ is non-decreasing and such that $f(k+1) - f(k) \leq 1$ for all $k \in [m-1]$.
\end{remark}

\section{The Case of Rank Loss}
\label{sec:rankloss}

In the case of the rank loss, we assume predictions in the form of rankings instead of labelings. Ignoring the possibility of ties, such a ranking can be represented in the form of a permutation $\pi$ of $[m]$, where $\pi(i)$ is the index $j$ of the label $\lambda_j$ on position $i$ (and $\pi^{-1}(j)$ the position of label $\lambda_j$). The rank loss then counts the number of incorrectly ordered label-pairs, that is, the number of pairs $\lambda_i, \lambda_j$ such that $\lambda_i$ is ranked worse than $\lambda_j$ although $\lambda_i$ is relevant while $\lambda_j$ is irrelevant:
\begin{equation*}
\ell_R(\vec{y} , \pi) = \sum_{(i,j): y_i > y_j} \left\llbracket \pi^{-1}(i) > \pi^{-1}(j) \right\rrbracket  \, ,
\end{equation*}
or equivalently, 
\begin{equation}\label{eq:rankl}
\ell_R(\vec{y} , \pi) = \sum_{1 \leq i < j \leq m} \llbracket y_{\pi(i)} = 0 \wedge y_{\pi(j)} = 1 \rrbracket \, .
\end{equation}
Thus, given that the ground-truth labeling is distributed according to the probability $\Prob(\cdot \given \vec{x})$, the expected loss of a ranking $\pi$ is
\begin{align}\label{eq:epi}
\mathbf{E}(\pi)  \defi \mathbf{E}\left(\ell_R(\vec{y}, \pi)\right)  = \sum_{1 \leq i < j \leq m} \Prob_{\pi(i) , \pi(j)}(0,1 \given \vec{x}) \, ,
\end{align}
where $\Prob_{u,v}$ is the pairwise marginal 
\begin{align}\label{eq:pairm}
\Prob_{u,v}(a,b \given \vec{x}) = \sum_{\vec{y} \in \mathcal{Y}: y_u = a, y_v = b} \Prob(\vec{y}  \given \vec{x}) \, .
\end{align}
In the following, we first recall the risk-minimizer for the rank loss as introduced above and then generalize it to the case of partial predictions. To simplify notation, we omit the dependence of probabilities on $\vec{x}$ (for example, we write $\Prob_{u,v}(a,b)$ instead of $\Prob_{u,v}(a,b \given \vec{x})$), and write $(i)$ as indices of permuted labels instead of $\pi(i)$. 
We also use the following notation: For a labeling $\vec{y}$, let $r(\vec{y})= \sum_{i=1}^m y_i$ be the number of relevant labels, and $c(\vec{y}) = r(\vec{y})(m-r(\vec{y}))$ the number of relevant/irrelevant label pairs (and hence an upper bound on the rank loss). 

A risk-minimizing ranking $\pi$, i.e., a ranking minimizing (\ref{eq:epi}), is provably obtained by sorting the labels $\lambda_i$ in decreasing order of the probabilities $p_i = \Prob_i(1 \given \vec{x})$, i.e., according to their probability of being relevant \citep{dembczynski:2012}. Thus, an optimal prediction $\pi$ is such that
\begin{equation}\label{eq:sort}
p_{(1)} \geq p_{(2)} \geq \cdots \geq p_{(m)} \,.
\end{equation}
To show this result, let $\bar{\pi}$ denote the reversal of $\pi$, i.e., the ranking that reverses the order of the labels. Then, for each pair $(i,j)$ such that $y_i > y_j$, either $\pi$ or $\bar{\pi}$ incurs an error, but not both. Therefore, $\ell_R(\vec{y} , \pi) + \ell_R(\vec{y} , \bar{\pi}) = c(\vec{y})$, and 
\begin{equation}\label{eq:qay}
\ell_R(\vec{y} , \pi) - \ell_R(\vec{y} , \bar{\pi}) = 2 \ell_R(\vec{y} , \pi) - c(\vec{y}) \, .
\end{equation}
Since $c(\vec{y})$ is a constant that does not depend on $\pi$, minimizing $\ell_R(\vec{y} , \pi)$ (in expectation) is equivalent to minimizing the difference $\ell_R(\vec{y} , \pi) - \ell_R(\vec{y} , \bar{\pi})$. For the latter, the expectation (\ref{eq:epi}) becomes 
\begin{align} \label{eq:acv}
\mathbf{E}'(\pi) & = \sum_{1 \leq i < j \leq m} \Prob_{(i) , (j)}(0,1) -  \Prob_{(i) , (j)}(1,0) \\
& = \sum_{1 \leq i < j \leq m} p_{(j)} -  p_{(i)}  \nonumber \\
&= \sum_{1 \leq i \leq m} (2 i - (m + 1))  p_{(i)}   \, , \nonumber 
\end{align}
where the transition from the first to the second sum is valid because \citep{dembczynski:2010}
\begin{align*}
& \Prob_{u,v}(0,1) -  \Prob_{u,v}(1,0)   \nonumber \\
&= \Prob_{u,v}(0,1) + \Prob_{u,v}(1,1)  - \Prob_{u,v}(1,1) -  \Prob_{u,v}(1,0) \\
  & = \Prob_v(1) - \Prob_u(1)  = p_{v} -  p_{u} \, . \nonumber
\end{align*}
From \eqref{eq:acv}, it is clear that a risk-minimizing ranking $\pi$ is defined by \eqref{eq:sort}.

To generalize this result, let us look at the rank loss of a partial prediction of size $d \in [m]$, i.e., a ranking of a subset of $d$ labels. To simplify notation, we identify such a prediction, not with the original set of indices of the labels, but the positions of the corresponding labels in the sorting (\ref{eq:sort}). Thus, a partial prediction of size $d$ is identified by a set of indices $K = \{ k_1, \ldots , k_d \}$ such that $k_1 < k_2 < \cdots < k_d$, where $k \in K$ means that the label $\lambda_{(k)}$ with the $k$th largest probability $p_{(k)}$ in (\ref{eq:sort}) is included. According to the above result, the optimal ranking $\pi_K$ on these labels is the identity, and the expected loss of this ranking is given by
\begin{equation}\label{eq:elpp}
\mathbf{E}(\pi_K) = \sum_{1 \leq i < j \leq d} \Prob_{(k_i) , (k_j)} (0,1) \,  .
\end{equation}

\begin{lemma}\label{pro:lpp}
Assuming (conditional) independence of label probabilities in the sense that $\Prob_{i , j} (y_i,y_j) = \Prob_{i} (y_i) \Prob_{j} (y_j)$, the generalized loss (\ref{eq:glosscount}) is minimized in expectation by a partial prediction with decision set of the form 
\begin{equation}\label{eq:dsel}
K_d = \lAngle a,b \rAngle \defi \{1, \ldots , a\} \cup \{b , \ldots , m\} \, ,
\end{equation}
with $1 \leq a < b \leq m$ and $m+a-b+1=d$.
\end{lemma}

According to the previous lemma, an optimal $d$-selection $K_d$ leading to an optimal (partial) ranking of length $d$ is always a ``boundary set'' of positions in the ranking (\ref{eq:sort}). The next lemma establishes an important relationship between optimal selections of increasing length. 

\begin{lemma}\label{pro:lpp2}
Let $K_d = \lAngle a,b \rAngle$ be an optimal $d$-selection (\ref{eq:dsel}) for $d \geq 2$. At least one of the extensions $\lAngle a+1,b \rAngle$ or $\lAngle a,b-1 \rAngle$ of $K_d$ is an optimal $(d+1)$-selection.
\end{lemma}

Thanks to the previous lemma, a risk-minimizing partial ranking can be constructed quite easily (in time $O(m \log(m))$. First, the labels are sorted according to (\ref{eq:sort}). Then, an optimal decision set is produced by starting with the boundary set $\lAngle 1,m \rAngle$ and increasing this set in a greedy manner ( a concrete algorithm is given in the supplementary material). 

\begin{proposition}\label{pro:minimize_rankgl}
Given a query instance $\vec{x}$, assume conditional probabilities $\Prob(y_i = 1 \given \vec{x}) = h_i(\vec{x})$ are made available by an MLC predictor $\mathbf{h}$. A risk-minimizing partial ranking can be constructed in time $O(m \log(m))$. 
\end{proposition}

\begin{remark}\label{pro:rankcal}
The extension \eqref{eq:glosscount} of the rank loss is not uncertainty-aligned.
\end{remark}

Since a prediction is a (partial) ranking instead of a (partial) labeling, the property of monotonicity as defined in Section \ref{sec:losspr} does not apply in the case of rank loss. Although it would be possible to generalize this property, for example by looking at (in)correctly sorted label pairs instead of (in)correct labels, we refrain from a closer analysis here. 

\section{The Case of F-measure}
\label{sec:Fmeasure}

The F-measure is the harmonic mean of precision and recall and can be expressed as follows:
\begin{align}\label{eq:Fmeasure}
F(\vec{y}, \hat{\vec{y}}) \defi \frac{2\sum_{i=1}^m y_i \hat{y}_i}{\sum_{i=1}^m( y_i + \hat{y}_i)}\, .
\end{align}
The problem of finding the expected F-maximizer 
\begin{align}\label{eq:Flm}
\hat{\vec{y}} &= \arg\max_{\hat{\vec{y}} \in \mathcal{Y}} \mathbf{E} \left(F(\vec{y}, \hat{\vec{y}}) \right) \\
& = \arg\max_{\hat{\vec{y}} \in \mathcal{Y}} \sum_{\vec{y} \in \mathcal{Y}} F(\vec{y}, \hat{\vec{y}}) \cdot \Prob(\vec{y} \given \vec{x}) \nonumber
\end{align}
has been studied quite extensively in the literature \citep{chai:2005,dembczynski:2011,decubber:2018,jansche:2007,lewis:1995,quevedo:2012,waegeman:2014,ye:2012}. Obviously, the optimization problem \eqref{eq:Flm} can be decomposed into an inner and an outer maximization as follows:
\begin{align}
\hat{\vec{y}}^k & \defi  \arg\max_{\hat{\vec{y}} \in \mathcal{Y}_k} \mathbf{E} \left(F(\vec{y}, \hat{\vec{y}}) \right) \, , \label{eq:inner}\\
\hat{\vec{y}} & \defi  \arg\max_{k \in \{ 0, \ldots , m \}} 
\mathbf{E} \left( F(\vec{y}, \hat{\vec{y}}^k)  \right)  
\, , \label{eq:outer}
\end{align}
where $\mathcal{Y}_k \defi \{\hat{\vec{y}} \in \mathcal{Y} \vert \sum_{i = 1}^m \hat{y}_i =k\}$ denotes the set of all predictions with exactly $k$ positive labels.

\cite{lewis:1995} showed that, under the assumption of conditional independence, the F-maximizer has always a specific form: it predicts the $k$ labels with the highest marginal probabilities $p_i$ as relevant, and the other $m-k$ labels as irrelevant. More specifically, for any number $k= 0, \ldots,m$, the solution of the optimization problem \eqref{eq:inner}, namely a $k$-optimal solution $\hat{\vec{y}}^k$, is obtained by setting $\hat{y}_i = 1$ for the $k$ labels with the highest marginal probabilities $p_i$, and $\hat{y}_i = 0$ for the remaining ones. Thus, the F-maximizer \eqref{eq:outer} can be found as follows:
\begin{itemize}
\item The labels $\lambda_i$ are sorted in decreasing order of their (predicted) probabilities $p_i$.  
\item For every $k \in \{ 0, \ldots , m \}$, the optimal prediction $\hat{\vec{y}}^k$ is defined as described above. 
\item For each of these $\hat{\vec{y}}^k$, the expected F-measure is computed. 
\item As an F-measure maximizer $\hat{\vec{y}}$, the $k$-optimal prediction $\hat{\vec{y}}^k$ with the highest expected F-measure is adopted. 
\end{itemize}
Overall, the computation of $\hat{\vec{y}}$ can be done in time $O(m^2)$ \citep{decubber:2018,ye:2012}.

To define the generalization of the F-measure, we first turn it into the loss function $\ell_F(\vec{y}, \hat{\vec{y}}) \defi 1 - F(\vec{y}, \hat{\vec{y}})$. The generalized loss is then given by 
\begin{align*}
L_F(\vec{y}, \hat{\vec{y}}) \defi 1 - F(\vec{y}_D, \hat{\vec{y}}_D)  +  f(|A(\hat{\vec{y}})|).
\end{align*}
Minimizing the expectation of this loss is obviously equivalent to maximizing the following generalized F-measure in expectation:
\begin{align}\label{eq:glF}
F_G(\vec{y}, \hat{\vec{y}}) \defi 
\begin{cases} 
1 -  f (a)  & \text{ if } a = m  \, ,\\ 
\frac{2\sum_{i \in D(\hat{\vec{y}})} y_i \hat{y}_i}{\sum_{i \in D(\hat{\vec{y}})} (y_i + \hat{y}_i)} -  f (a) &  \text{ otherwise,}
\end{cases} 
\end{align}
where $a \defi |A(\hat{\vec{y}})|$.

\begin{remark}\label{pro:1F}
If $f$ in \eqref{eq:glF} is a strictly increasing function, then 
\begin{itemize}
\item[-] turning an incorrect prediction or an abstention on a label $\lambda_i$ into a correct prediction increases the generalized F-measure, whereas
\item[-] turning an incorrect prediction into an abstention may decrease the measure.
\end{itemize} 
Therefore, the generalized F-measure \eqref{eq:glF} is not monotonic.
\end{remark}

The F-maximizer $\hat{\vec{y}}$ of the generalized F-measure \eqref{eq:glF} is given by
\begin{align}\label{eq:FlmG}
\hat{\vec{y}} &= \arg\max_{\hat{\vec{y}} \in \mathcal{Y}_{pa}} \mathbf{E} \left(F_G(\vec{y}, \hat{\vec{y}}) \right) \\
& = \arg\max_{\hat{\vec{y}} \in \mathcal{Y}_{pa}} \sum_{\vec{y} \in \mathcal{Y}} F(\vec{y}_D, \hat{\vec{y}}_D) \cdot \Prob(\vec{y} \given \vec{x}) - f(a) \, . \nonumber
\end{align}

In the following, we show that the F-maximizer of the generalized F-measure \eqref{eq:glF} can be found in the time $O(m^3)$. For any $k = 0, \ldots,m$, denote by 
\begin{align}
\mathcal{Y}_{pa}^k \defi \left\{\hat{\vec{y}} \in \mathcal{Y}_{pa} \, \Big\vert \sum_{i\in D(\hat{\vec{y}})} \hat{y}_i = k  \right\}\, .
\end{align}
The optimization problem \eqref{eq:FlmG} is decomposed into an inner and an outer maximization  as~follows:
\begin{align}
\hat{\vec{y}}^k & \defi  \arg\max_{\hat{\vec{y}} \in \mathcal{Y}_{pa}^k } \mathbf{E} \left(F_G(\vec{y}, \hat{\vec{y}}) \right) \, , \label{eq:innerG}\\
\hat{\vec{y}} & \defi  \arg\max_{\hat{\vec{y}} \in \{\hat{\vec{y}}^k \vert k = 0, \ldots,m \}} \mathbf{E} \left(F_G(\vec{y}, \hat{\vec{y}}^k) \right) \, . \label{eq:outerG}
\end{align}

\begin{lemma} \label{lem:F2}
For any partial prediction $\hat{\vec{y}} \in \mathcal{Y}_{pa}$ and any index $j \in D(\hat{\vec{y}})$, 
\begin{itemize}
\item[-] $\mathbf{E} \left(F_G(\vec{y}_D, \hat{\vec{y}}_D) \right)$ is an increasing function of $p_j$ if $\hat{y}_j = 1$;
\item[-] $\mathbf{E} \left(F_G(\vec{y}_D, \hat{\vec{y}}_D) \right)$ is a decreasing function of $p_j$ if $\hat{y}_j = 0$.
\end{itemize}
\end{lemma}

\begin{lemma}\label{lem:F3}
Let $\pi$ be the permutation that sorts the labels in decreasing order of the marginal probability $\Prob_i(y_i \given \vec{x})$ defined in \eqref{eq:sort}. Assuming (conditional) independence of label probabilities in the sense that    
\begin{align}\label{eq:independence}
\Prob(\vec{y} \vert \vec{x}) =\prod_{i=1}^m p_i^{y_i}(1-p_i)^{1-y_i} \, ,
\end{align}
the generalized F-measure (\ref{eq:glF}) is maximized in expectation by an optimal $k$-prediction $\hat{\vec{y}}^k$ with decision set of the form 
\begin{equation}\label{eq:dselF}
D(\hat{\vec{y}}^k) = \lAngle k,l \rAngle \defi \{1, \ldots , k\} \cup \{l , \ldots , m\} \, ,
\end{equation}
with some $l \geq k+1$ and 
   \begin{align} \label{eq:optimal_prediction}
   \hat{y}_{(i)} = \begin{cases}
   \quad 1 & \text{ if } \,  i \in \{ 1,\ldots, k\} \, , \\
   \quad 0 & \text{ if } \, i \in \{ l, \ldots, m \} \, .  \\
   \end{cases} 
   \end{align}
\end{lemma}
Thanks to the previous lemma, a maximizer $\hat{\vec{y}}$ of the generalized F-measure \eqref{eq:glF} can be constructed following a procedure similar to the case of the rank loss. First, the labels are sorted according to \eqref{eq:sort}. Then, we evaluate all possible partial predictions with decision sets $\lAngle k,l \rAngle$ of the form \eqref{eq:dselF}, and find the one with the highest expected F-measure \eqref{eq:glF} (a concrete algorithm is given in the supplementary material). 

\begin{proposition} \label{pro:maximize_F}
Given a query instance $\vec{x}$, assume conditional probabilities $p_i =  \Prob(y_i = 1 \given \vec{x}) = h_i(\vec{x})$ are made available by an MLC predictor $\mathbf{h}$. Assuming (conditional) independence of label probabilities in the sense of \eqref{eq:independence}, a prediction $\hat{\vec{y}}$ maximizing the generalized F-measure \eqref{eq:glF} in expectation is constructed in time $O(m^3)$. 
\end{proposition}

\section{Related Work}
\label{sec:rlw}

In spite of extensive research on multilabel classification in the recent past, there is surprisingly little work on abstention in MLC. A notable exception is an approach by \citet{pillai:2013}, who focus on the F-measure as a performance metric. They tackle the problem of maximizing the F-measure on a subset of label predictions, subject to the constraint that the effort for manually providing the remaining labels (those on which the learner abstains) does not exceed a pre-defined value. The decision whether or not to abstain on a label is guided by two thresholds on the predicted degree of relevance, which are tuned in a suitable manner. Even though this is an interesting approach, it is arguably less principled than ours, in which optimal predictions are derived in a systematic way, based on decision-theoretic ideas and the notion of Bayes-optimality. Besides, \cite{pillai:2013} offer a solution for a specific setting but not a general framework for MLC with partial abstention.   

More indirectly related is the work by \citet{park}, who investigate the uncertainty in multilabel classification. They propose a modification of the entropy measure to quantify the uncertainty of an MLC prediction. Moreover, they show that this measure correlates with the accuracy of the prediction, and conclude that it could be used as a measure of acceptance (and hence rejection) of a prediction. While \citet{park} focus on the uncertainty of a complete labeling $\vec{y}$, \citet{destercke:2015} and \citet{antonucci:2017} quantify the uncertainty in individual predictions $y_i$ using imprecise probabilities and so-called credal classifiers, respectively. Again, corresponding estimates can be used for the purpose of producing more informed decisions, including partial~predictions.

\section{Experiments}
\label{sec:exp}

In this section, we present an empirical analysis that is meant to show the effectiveness of our approach to prediction with abstention. To this end, we perform experiments on a set of standard benchmark data sets from the MULAN repository\footnote{http://mulan.sourceforge.net/datasets.html} (cf.\ Table \ref{tab:datasets}), following a $10$-fold cross-validation procedure. 

\subsection{Experimental Setting}

For training an MLC classifier, we use binary relevance (BR) learning with logistic regression (LR) as base learner (in its default setting in sklearn, i.e., with regularisation parameter set to $1$)\footnote{For an implementation in Python, see \url{http://scikit.ml/api/skmultilearn.base.problem_transformation.html}.}.
Of course, more sophisticated techniques could be applied, and results using classifier chains are given in the supplementary material. However, since we are dealing with decomposable losses, BR is well justified. Besides, we are first of all interested in analyzing the effectiveness of abstention, and less in maximizing overall performance. All competitors essentially only differ in how the conditional probabilities provided by LR are turned into a (partial) MLC prediction.

\begin{table}\caption{Data sets used in the experiments}
\label{tab:datasets}
\begin{center}
\begin{sc}
\begin{tabular}{lcccr}
\toprule
\# &name    &  \# inst. & \# feat. & \# lab. \\
\midrule
1 &cal500     &502  &68   &174   \\
2 &emotions  &593  &72   &6  \\
3 &scene     &2407 &294  &6  \\ 
4 &yeast     &2417 &103  &14 \\
5 &mediamill  &43907  &120    &101 \\
6 &nus-wide   &269648 &128    &81 \\
\bottomrule
\end{tabular}
\end{sc}
\end{center}
\end{table}


We first compare the performance of reliable classifiers to the conventional BR classifier that makes full predictions ({\sc MLC}) as well as the cost of full abstention ({\sc ABS})---these two serve as baselines that MLC with abstention should be able to improve on. A classifier is obtained as a risk-minimizer of the extension \eqref{eq:glosscount} of Hamming loss \eqref{eq:hamming}, instantiated by the penalty function $f$ and the constant $c$. We conduct a first series of experiments (SEP) with linear penalty $f_1(a) = a \cdot c$, where $c \in [0.05,0.5]$, and a second series (PAR) with concave penalty $f_2(a) = (a\cdot m\cdot c)/(m+a)$, varying $c \in [0.1,1]$.  The performance of a classifier is evaluated in terms of the average loss. Besides, we also compute the average abstention size $|A(\hat{\vec{y}})|/m$. 

The same type of experiment is conducted for the rank loss (with {\sc MLC} and {\sc ABS} denoting full prediction and full abstention, respectively). A predicted ranking is a risk-minimizer of the extension \eqref{eq:glosscount} instantiated by the penalty function $f$ and  the constant $c$. We conduct a first series of experiments (SEP) with $f_1$ as above and $c \in [0.1,1]$, and a second series (PAR) with $f_2$ as above and $c \in [0.2,2]$.

\subsection{Results}

\begin{figure}[!t]
\centering 
\includegraphics[width=.95\columnwidth]{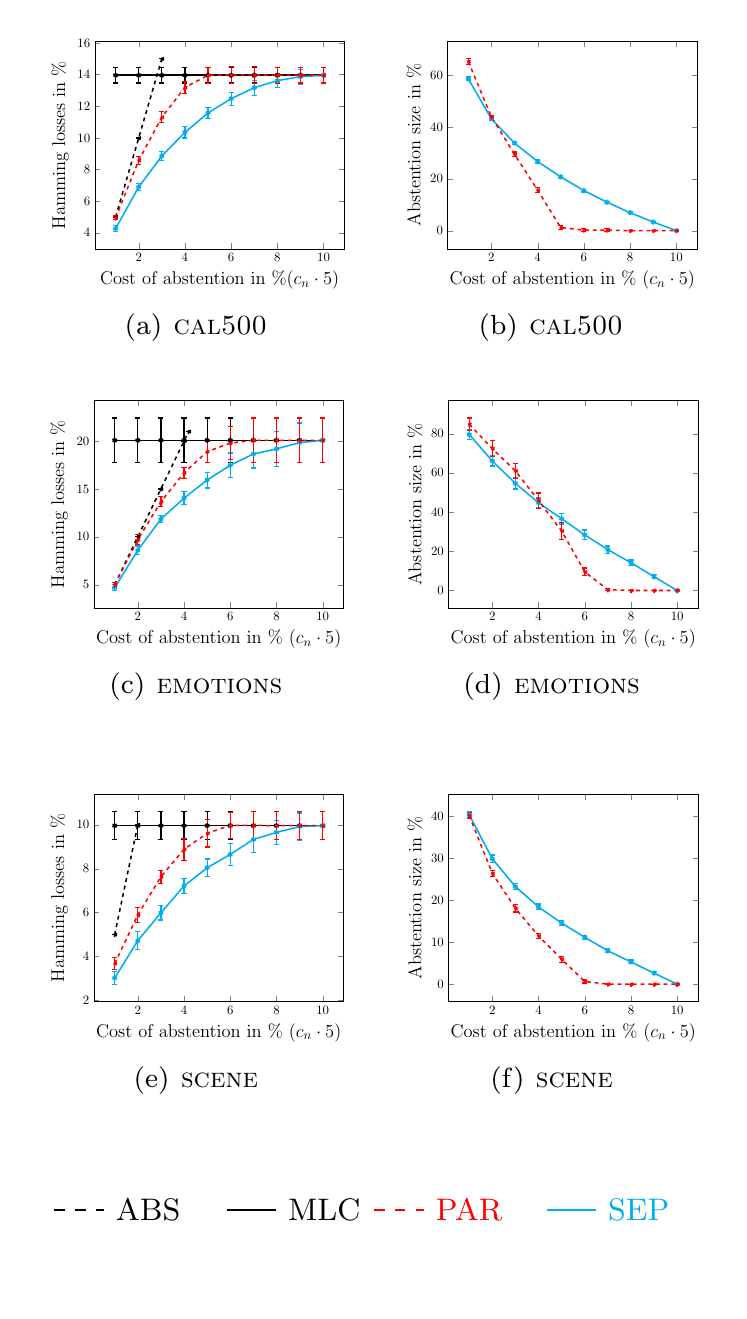}
\caption{Binary relevance with logistic regression: Experimental results in terms of expected Hamming loss $(L_H \cdot 100)/m$ and abstention size (in percent) for $f_1(a) = a \cdot c$ ({\sc SEP}) and $f_2(a) = (a\cdot m\cdot c)/(m+a)$ ({\sc PAR}), as a function of the cost of abstention.}
\label{fig:short_br_lr_h}  
\end{figure}

\begin{figure}[!t]
\centering 
\includegraphics[width=.95\columnwidth]{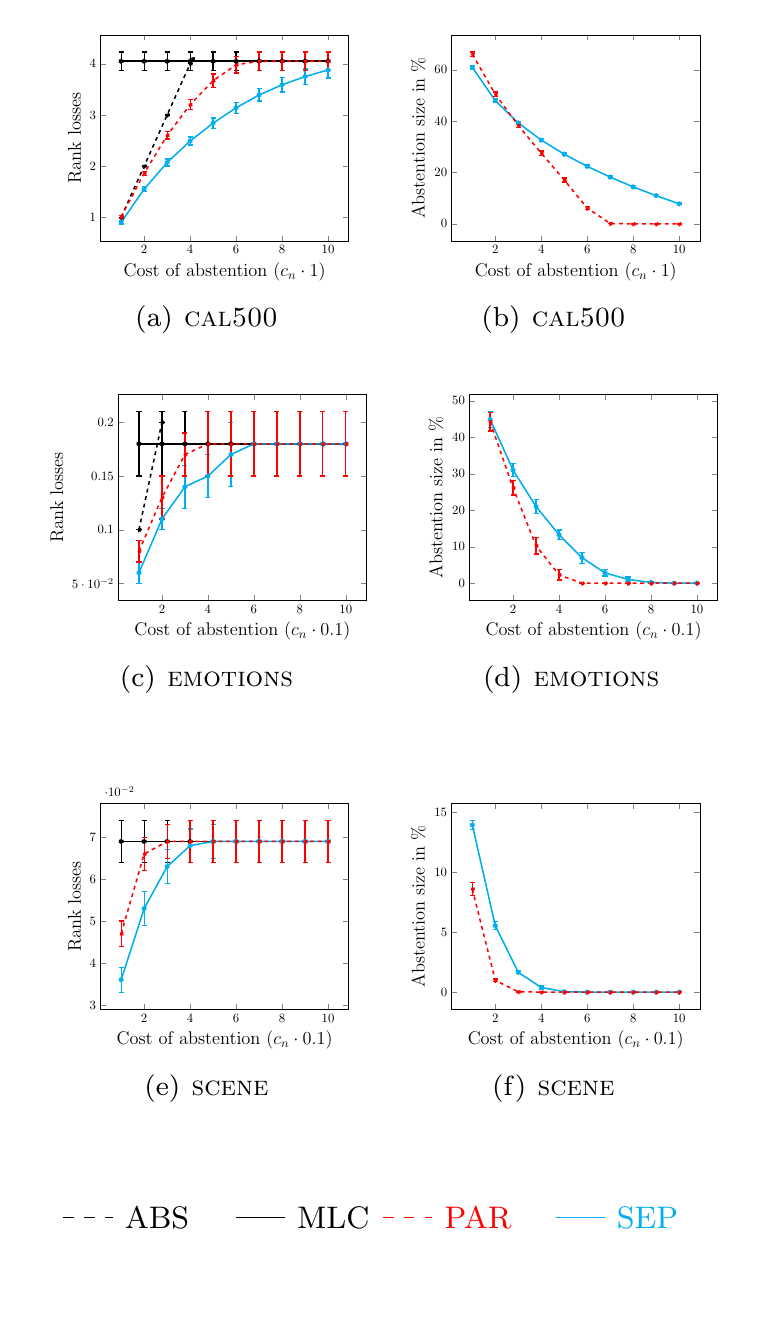}
\caption{Binary relevance with logistic regression: Experimental results in terms of expected rank loss $L_R/m$ and abstention size (in percent) for $f_1(a) = a \cdot c$ ({\sc SEP}) and $f_2(a) = (a\cdot m\cdot c)/(m+a)$ ({\sc PAR}), as a function of the cost of abstention.}
\label{fig:short_br_lr_r}  
\end{figure}

The results (illustrated in Figures \ref{fig:short_br_lr_h} and \ref{fig:short_br_lr_r} for three data sets--results for the other data sets are similar and can be found in the supplementary material) clearly confirm our expectations. The Hamming loss (cf.\ Figure \ref{fig:short_br_lr_h}) under partial abstention is often much lower than the loss under full prediction and full abstention, showing the effectiveness of the approach. When the cost $c$ increases, the loss increases while the abstention size decreases, with a convergence of the performance of SEP and PAR to the one of {\sc MLC} at $c = 0.5$ and $c = 1$, respectively. 

Similar results are obtained in the case of rank loss (cf.\ Figure \ref{fig:short_br_lr_r}), except that convergence to the performance of {\sc MLC} is slower (i.e., requires lager cost values $c$, especially on the data set {\sc cal500}). This is plausible, because the cost of a wrong prediction on a single label can be as high as $m-1$, compared to only 1 in the case of Hamming.

Due to space restrictions, we transferred experimental results for the generalized F-measure to the supplementary material. These results are very similar to those for the rank loss. In light of the observation that the respective risk-minimizers have the same structure, this is not very surprising. 


The supplementary material also contains results for other MLC algorithms, including BR with support vector machines (using Platt-scaling \citep{lin:2007,platt:1999} to turn scores into probabilities) as base learners and classifier chains \citep{read:2009} with LR and SVMs as base learners. Again, the results are very similar to those presented above. 

\section{Conclusion}
\label{sec:con}

This paper presents a formal framework of MLC with partial abstention, which builds on two main building blocks: First, the extension of an underlying MLC loss function so as to accommodate abstention in a proper way, and second the problem of optimal prediction, that is, minimizing this loss in expectation. 

We instantiated our framework for the Hamming loss, the rank loss, and the F-measure, which are three important and commonly used loss functions in multi-label classification. We elaborated on properties of risk-minimizers, showed them to have a specific structure, and devised efficient methods to produce optimal predictions. Experimentally, we showed these methods to be effective in the sense of reducing loss when being allowed to abstain.       

In future work, we will further elaborate on our formal framework. As a concrete next step, we plan to investigate instantiations for other loss functions commonly used in MLC and the cases of label dependence \citep{dembczynski:2012,waegeman:2014}. 

\section*{Acknowledgments}
We thank Willem Waegeman for interesting discussions and useful suggestions. This work was supported by the Germany Research Foundation (DFG) (grant numbers HU 1284/19).

\medskip
\begin{quote}
\begin{small}

\end{small}
\end{quote}

\newpage
\clearpage


\begin{appendix}

\begin{center}
\LARGE \textbf{Supplementary Material}
\end{center}

\section{Proof of Proposition \ref{pro:decloss}}
Denote by $\ell^i_{0,1}:= \ell_i(0,1)$ and $\ell^i_{1,0} := \ell_i(1,0)$, 
\begin{equation*}
s_i= \min_{\hat{y}_i \in \{0,1\}}\mathbf{E}(\ell_i(y_i,\hat{y}_i))= \min\left(\ell^i_{0,1} \cdot p_i,\ell^i_{1,0} \cdot (1-p_i)\right)\,  ,
\end{equation*}
and $\pi$ is the permutation sorts the labels in increasing order of the label-wise expected losses, i.e., $s_{\pi(1)} \leq \cdots \leq s_{\pi(m)}$.

The expected loss of the extension \eqref{eq:glosscount} can be expressed~as
\begin{align*} 
\mathbf{E} &\left(L(\vec{y}, \hat{\vec{y}}) \right) = \sum_{\vec{y} \in \mathcal{Y}} L(\vec{y} , \hat{\vec{y}}) \cdot \Prob(\vec{y} \given \vec{x}) \\
=& \sum_{\vec{y} \in \mathcal{Y}}\left( \sum_{i \in D(\hat{\vec{y}})} \ell_i(y_i , \hat{y_i} ) + f(|A(\hat{\vec{y}})|)\right)\cdot \Prob(\vec{y} \given \vec{x}) \\
 =& \sum_{\substack{ i : \hat{y}_i = 0 \\ i \in D(\hat{\vec{y}})}} \ell^i_{0,1} \cdot p_i  +  \sum_{\substack{ i : \hat{y}_i = 1 \\ i \in D(\hat{\vec{y}})}} \ell^i_{1,0} \cdot (1-p_i) + f(|A(\hat{\vec{y}})|)\, .
\end{align*}
The problem of loss minimization can be rewritten as 
\begin{align*}
\hat{\vec{y}} & \in  \operatorname*{argmin}_{d\in [m]} \left(\operatorname*{argmin}_{\substack{\vec{y} \in \mathcal{Y} \\|D(\hat{\vec{y}})| = d}} \mathbf{E} \left(L(\vec{y}, \hat{\vec{y}}) \right)\right) \\
&= \operatorname*{argmin}_{d\in [m]}  \mathbf{E} \left(L(\vec{y},\hat{\vec{y}}_d) \right) \,,
\end{align*} 
where
\begin{align*}
\hat{\vec{y}}_d & \in  \operatorname*{argmin}_{\substack{\hat{\vec{y}} \in \mathcal{Y} \\|D(\hat{\vec{y}})| = d}} \mathbf{E} \left(L(\vec{y},\hat{\vec{y}}) \right)\, \\
&= \operatorname*{argmin}_{\substack{\hat{\vec{y}} \in \mathcal{Y}\\|D(\hat{\vec{y}})| = d}}\left(\sum_{\substack{ i : \hat{y}_i = 0 \\ i \in D(\hat{\vec{y}})}} \ell^i_{0,1} \cdot p_i  + \sum_{\substack{ i : \hat{y}_i = 1 \\ i \in D(\hat{\vec{y}})}} \ell^i_{1,0} \cdot (1-p_i)\right) \\
&= \operatorname*{argmin}_{\substack{\hat{\vec{y}} \in \mathcal{Y}\\|D(\hat{\vec{y}})| = d}}\sum_{ i \in D(\hat{\vec{y}})} \min\left(\ell^i_{0,1} \cdot p_i, \ell^i_{1,0} \cdot (1-p_i) \right)\, ,\\
& = \operatorname*{argmin}_{\substack{\hat{\vec{y}} \in \mathcal{Y}\\|D(\hat{\vec{y}})| = d}}\sum_{ i \in D(\hat{\vec{y}})} s_i\, .
\end{align*}
The prediction $\hat{\vec{y}}_d$, $d\in [m]$, is specified by the index set 
\begin{equation*}
D_d(\hat{\vec{y}}_d) =  \{\pi(1), \ldots , \pi(d)\} \, ,
\end{equation*}
because replacing any $i \in D_d(\hat{\vec{y}}_d) $ by $i' \in [m]\setminus D_d(\hat{\vec{y}}_d)$ clearly increases $s_i$. 

\section{Proof of Corollary \ref{pro:hamopt}}
The proof is obvious because in the case of Hamming loss, the degrees of uncertainty 
\begin{align*}
u_i &= 2 \min(p_i,1-p_i)  \\
& = 2 \min_{\hat{y}_i \in \{0,1\}}\mathbf{E}(\ell_i(y_i, \hat{y}_i)).
\end{align*}
Thus, sorting the labels in increasing order of the degrees of uncertainty $u_i$ \eqref{eq:unc} is equivalent to doing so with the label-wise expected losses $\min_{\hat{y}_i \in \{0,1\}}\mathbf{E}(\ell_i(y_i, \hat{y}_i))$. To this end, the proof of Corollary \ref{pro:hamopt} is carried out consequently from the proof of Proposition \ref{pro:decloss}. 

\section{Proof of Corollary \ref{pro:Hamming1}}
It is easy to verify that the extension \eqref{eq:glosscount} of the Hamming loss is uncertainty-aligned since its risk-minimizers are always of the form~\eqref{eq:hamsl}. 

In the following, we show that the risk-minimizer of the generalized Hamming loss \eqref{eq:gloss1} can be found simply by abstaining those labels with $\min(p_i, 1 - p_i) > c$. 

The expected loss of the generalized Hamming loss \eqref{eq:gloss1} is
\begin{align*} 
\mathbf{E} \left(L_H(\vec{y},\hat{\vec{y}}) \right) & = \sum_{\vec{y} \in \mathcal{Y}} L_H(\vec{y} , \hat{\vec{y}}) \cdot \Prob(\vec{y} \given \vec{x}) \\
& =  \sum_{\substack{ i : \hat{y}_i = 0 \\ i \in D(\hat{\vec{y}})}} p_i  +  \sum_{\substack{ i : \hat{y}_i = 1 \\ i \in D(\hat{\vec{y}})}} (1-p_i) + |A(\hat{\vec{y}})|\cdot c \\
& = \sum_{\substack{ i : \hat{y}_i = 0 \\ i \in D(\hat{\vec{y}})}} p_i  +  \sum_{\substack{ i : \hat{y}_i = 1 \\ i \in D(\hat{\vec{y}})}} (1-p_i)+ \sum_{i \in A(\hat{\vec{y}})}c \,.\\
\end{align*}
Finding the risk-minimizer is thus equivalent to solving the following optimization problem:
\begin{align*} 
\hat{\vec{y}}&= \operatorname*{argmin}_{\vec{y} \in \mathcal{Y}_{pa}} \left(\sum_{i : \hat{y}_i = 0} p_i  +  \sum_{ i : \hat{y}_i = 1 } (1-p_i)+ \sum_{i : \hat{y}_i = \bot}c \right) \,.
\end{align*}
Thus to minimize the expected loss, we should abstain all the index $i \in [m]$ s.t $c < \min(p_i,1-p_i)$ and return an optimal $d$-prediction $D(\hat{\vec{y}}):= \{ i | c > \min(p_i,1-p_i)\}$.

The proof of Corollary \ref{pro:Hamming1} is completed by rearranging the elements of $D(\hat{\vec{y}})$ according the increasing order of the degrees of uncertainties $u_i$ \eqref{eq:unc}, $i \in D(\hat{\vec{y}})$.

\section{Proof of Remark \ref{prop1}}

We start with the general setting that if $f:=f(|A(\hat{\vec{y}})|)$ and
\begin{equation}\label{eq:glosspro1}
f(k) - f(k-1) \in [0,1], \, \forall k \in [m] \, ,  
\end{equation}
the extension \eqref{eq:glosscount} of the Hamming loss \eqref{eq:hamming} is monotonic. 

Let us consider two predictions $\hat{\vec{y}}$ and $\hat{\vec{y}}'$, s.t, for a given $i \in [m]$, we have
\begin{align*}
\begin{cases}
\ell_H(y_i,\hat{y}_i) \prec  \ell_H(y_i,\hat{y}'_i), &\text{ and } \\
\ell_H(y_j,\hat{y}_j) = \ell_H(y_j,\hat{y}'_j), &\text{ if } j \neq i,
\end{cases}
\end{align*}
where $\ell_H(y_i,\hat{y}_i)$ can be: $\ell_w$ (a wrong prediction), $\ell_c$ (a correct prediction), and $\ell_a$ (an abstention). The preference relation $ \prec$ is defined s.t, $\ell_w \prec \ell_a \prec \ell_c$. 

We proceed by considering three possible combinations of $\left(\ell_H(y_i,\hat{y}_i), \ell_H(y_i,\hat{y}'_i) \right)$. For a seek of simplicity, let us denote by $k:=|A(\hat{\vec{y}})|$ the number of abstained labels in $\hat{y}$, then the number of abstention in $\hat{\vec{y}}'$ can be either $k' \in \{k-1, k,k+1\}$.

(i) $(\ell_w,\ell_c)$: in this case, we have $k' = k$ and $D(\hat{\vec{y}'})= D(\hat{\vec{y}})$. It is clear that $L(\vec{y},\hat{\vec{y}}) \geq L(\vec{y},\hat{\vec{y}}')$ since 
\begin{align*}
L_H(\vec{y},\hat{\vec{y}})  =& \sum_{\substack{j \in D(\hat{\vec{y}})\\j \neq i}}\ell_H(y_j, \hat{y}_j) + 1 + f(k) \\
 \geq & \sum_{\substack{j \in D(\hat{\vec{y}'})\\j \neq i}}\ell_H(y_j, \hat{y}_j) + f(k)\\
= & L_H(\vec{y},\hat{\vec{y}}') \, .
\end{align*}

(ii) $(\ell_w,\ell_a)$: in this case, we have $k' = k+1$ and $D(\hat{\vec{y}'})= D(\hat{\vec{y}})\setminus \{i\}$. We can easily validate that $L_H(\vec{y},\hat{\vec{y}}) \geq L_H(\vec{y},\hat{\vec{y}}')$ using the following analysis. Since $f(k+1) \leq 1 + f(k)$, thus
\begin{align*}
L_H(\vec{y},\hat{\vec{y}})  =& \sum_{\substack{j \in D(\hat{\vec{y}})\\j \neq i}}\ell_H(y_j, \hat{y}_j) + 1 + f(k)\\
 \geq& \sum_{j \in D(\hat{\vec{y}}')}\ell_H(y_j, \hat{y}_j) + f(k+1) \\
= & L_H(\vec{y},\hat{\vec{y}}') \, .
\end{align*} 

(iii) $(\ell_a,\ell_c)$: in this case, we have $k' = k-1$ and $D(\hat{\vec{y}'})\setminus \{i\}= D(\hat{\vec{y}})$. It is not difficult to see that $L_H(\vec{y},\hat{\vec{y}}) \geq L_H(\vec{y},\hat{\vec{y}}')$. Since $f(k) \geq f(k-1)$, thus
\begin{align*}
L_H(\vec{y},\hat{\vec{y}}) = & \sum_{j \in D(\hat{\vec{y}})}\ell_H(y_j, \hat{y}_j) + f(k)\\ 
 \geq & \sum_{\substack{j \in D(\hat{\vec{y}'})\\j \neq i}}\ell_H(y_i, \hat{y}_i) + 0 + f(k-1) \\
= &  L_H(\vec{y},\hat{\vec{y}}') \, .
\end{align*} 

\section{Proof of Lemma \ref{pro:lpp}}

Let $K = \{ k_1, \ldots , k_d \}$ specify a partial prediction of size $d$, and let $\vec{y}$ be the labeling restricted to the selected labels. Then
\begin{align*}
\mathbf{E} (c(\vec{y})) & = \mathbf{E} \left( \left( \sum_{1 \leq i \leq d} y_{k_i} \right) \left( d- \sum_{1 \leq i \leq d} y_{k_i} \right) \right) \\
& = \mathbf{E} \left( d \left( \sum_{1 \leq i \leq d} y_{k_i} \right) \right) - 
\mathbf{E} \left( \left( \sum_{1 \leq i \leq d} y_{k_i} \right)^2 \right) \\
& =   d \sum_{1 \leq i \leq d} \mathbf{E}( y_{k_i} ) - 
\sum_{1 \leq i,j \leq d} \mathbf{E} ( y_{k_i} y_{k_j} ) \\
& =   (d-1) \sum_{1 \leq i \leq d} \mathbf{E}( y_{k_i} ) - 
\sum_{1 \leq i \neq j \leq d} \mathbf{E} ( y_{k_i} y_{k_j} ) \\
& = (d-1)  \sum_{1 \leq i \leq d} p_{k_i} - \sum_{1 \leq i \neq j \leq d} p_{k_i}p_{k_j} \, ,
\end{align*}
where we exploited that $(y_i)^2 = y_i$ and the assumption of (conditional) independence as made in the proposition. 

According to (\ref{eq:qay}) and (\ref{eq:acv}), we can write the expected loss of a ranking $\pi_K$ 
\begin{align}
\mathbf{E}(\pi_K)  & =  \frac{1}{2} \mathbf{E} \left( \left( \ell_R (\vec{y}, \pi) - \ell_R(\vec{y}, \bar{\pi}) \right)  \right) + \frac{1}{2} \mathbf{E} (c(\vec{y})) \nonumber \\
& = \frac{1}{2} \sum_{1 \leq i \leq d} (2i - (d+1)) p_{k_i} \nonumber \\
& \quad +  \frac{d-1}{2}  \sum_{1 \leq i \leq d} p_{k_i} - \frac{1}{2} \sum_{1 \leq i \neq j \leq d} p_{k_i}p_{k_j} \nonumber \\
& =  \sum_{1 \leq i \leq d} (i - 1) p_{k_i} - \sum_{1 \leq i < j \leq d} p_{k_i}p_{k_j} \nonumber \\
& = \sum_{1 \leq i < j \leq d} p_{k_j} (1 - p_{k_i})  \, .
\label{eq:rf}
\end{align}
Next, we show that the expression (\ref{eq:rf}) is minimized by a selection of the form (\ref{eq:dsel}), i.e.,
\begin{equation*} 
K_d = \lAngle a,b \rAngle = \{1, \ldots , a\} \cup \{b , \ldots , m\} \, ,
\end{equation*}
as stated in the lemma. To this end, note that the derivative of (\ref{eq:rf}) with respect to $p_{k_u}$ is given by
$$
\delta_u = \sum_{i < u} (1 - p_{k_i}) - \sum_{j > u} p_{k_j} \, .
$$
Thus, recalling that $p_{(1)} \geq p_{(2)} \geq \cdots \geq p_{(m)}$, we can conclude that (\ref{eq:rf}) can be reduced (or at least kept equal) if, for some $u \in \{1,\ldots,d\}$, 
\begin{itemize}
\item[(i)] $\delta_u  \leq 0$ and $u-1 \not\in K_d$,
\item[(ii)] $\delta_u  \geq 0$ and $u+1 \not\in K_d$,
\end{itemize}
namely by replacing $u$ with $u-1$ in $K_d$ in case (i) and replacing $u$ with $u+1$ in case (ii). Let us call such a replacement a ``swap''.

Now, suppose that, contrary to the claim of the lemma, an optimal selection is not of the form (\ref{eq:dsel}) and cannot be improved by a swap either. Then we necessarily have a situation where $b_1, \ldots , b_u \in K_d$ is a block of consecutive indices such that $b_1-1 \not\in K$ and $b_u+1 \not\in K_d$. Moreover, let $a$ be the largest index in $K_d$ smaller than $b_1$ and $b$ the smallest index in $K_d$ bigger than $b_u$. Since a swap from $b_1$ to $b_1 -1$ is not valid, 
$$
\delta_{b_1}  =  \sum_{i \leq a} (1 - p_{k_i}) - \left( p_{b_2} + \ldots + p_{b_u} + \sum_{j \geq  b} p_{k_j} \right) > 0 \, .
$$
Likewise, since a swap from $b_u$ to $b_u +1$ is not valid,
$$
-\delta_{b_u} = - \sum_{i \leq a} (1 - p_{k_i}) -  \sum_{j=1}^{u-1} (1-p_{b_j})  +\sum_{j \geq  b} p_{k_j} > 0 \, .
$$
Summing up these two inequalities yields
$$
p_{b_1} - p_{b_u} > u-1 \, ,
$$
which is a contradiction. 

\section{Proof of Lemma \ref{pro:lpp2}}

We proceed under the assumption that $p_{(i)} \not \in \{0,1\}$, $\forall i \in [m]$.

Let $K_d = \lAngle a,b \rAngle$ be an optimal $d$-selection (\ref{eq:dsel}) for $d \geq 2$. Since $K_d$ is an optimal $d$-selection, neither a replacement from $a$ to $b-1$ nor a replacement from $b$ to $a+1$ on $K_d$ reduces the expected loss. Denote by $\delta_a^d$ and $\delta_b^d$ the derivative of $\mathbf{E}(\pi_{K_d})$ with respect to $p_b$ and $p_b$, thus,
\begin{align}
\delta_a^d &= \sum_{i \leq a-1} (1-p_{(i)}) - \sum_{b \leq j} p_{(j)} \leq 0 \, , \label{eq:claim1} \\
\delta_b^d &= \sum_{i \leq a} (1-p_{(i)}) - \sum_{b+1\leq j} p_{(j)} \geq 0 \, .  \label{eq:claim2}
\end{align}

Lemma \ref{pro:lpp} implies that there is an optimal $(d+1)$-selection $K_{d+1} = \lAngle l,r \rAngle$.  Denote by $\delta_l^{d+1}$ and $\delta_r^{d+1}$, the derivative of $\mathbf{E}(\pi_{K_{d+1}})$ with respect to $p_l$ and $p_r$, thus
\begin{align*}
\delta_l^{d+1} &= \sum_{i \leq l-1} (1-p_{(i)}) - \sum_{r \leq j} p_{(j)} \leq 0 \, ,  \\
\delta_r^{d+1} &= \sum_{i \leq l} (1-p_{(i)}) - \sum_{r+1\leq j} p_{(j)} \geq 0 \, .  
\end{align*}

Now, suppose that, contrary to the claim of the lemma, 
\begin{equation*}
\bigg( \lAngle l,r \rAngle \neq \lAngle a+1,b \rAngle
 \bigg)\wedge \bigg( \lAngle l,r \rAngle \neq  \lAngle a,b-1 \rAngle \bigg) \,.
\end{equation*}
Thus, $K_{d+1}$ has $2$ following possible forms: (i) $(a<l) \wedge (b<r)$ or (ii) $(l<a) \wedge (r< b)$. 

The proof of Lemma \ref{pro:lpp2} is completed by showing that both (i) and (ii) lead to the contradiction.

(i) $(a<l) \wedge (b<r)$: it is not difficult to see that
\begin{align*}
\sum_{i \leq a} (1-p_{(i)}) &\leq  \sum_{i \leq l-1} (1-p_{(i)}) \, , \\
- \sum_{b+1 \leq j} p_{(j)} &\leq - \sum_{r \leq j} p_{(j)} \, .
\end{align*}
Furthermore, the equality can not occur in both inequalities at the same time, otherwise 
\begin{align*}
&(a=l-1) \wedge (b+1 =r) \\
\Rightarrow & b-a = r - l \, . 
\end{align*}
Thus, $\delta_b^d < \delta_l^{d+1} \leq 0 $, that contradicts \eqref{eq:claim2}. 

(ii) $(l<a) \wedge (r< b)$: it is not difficult to see that
\begin{align*}
\sum_{i \leq a-1} (1-p_{(i)}) &\geq  \sum_{i \leq l} (1-p_{(i)}) \, , \\
- \sum_{b \leq j} p_{(j)}  &\geq - \sum_{r+1\leq j} p_{(j)} \, .
\end{align*}
Furthermore, the equality can not occur in both inequalities at the same time, otherwise 
\begin{align*}
&(a-1=l) \wedge (b =r+1) \\
\Rightarrow & b-a = r - l \, . 
\end{align*}
Thus, $\delta_a^d > \delta_r^{d+1} \geq 0$, that contradicts \eqref{eq:claim1}. 

\section{Proof of Proposition \ref{pro:minimize_rankgl}}
	\begin{algorithm} [!h]
	\caption{Expected rank loss minimization}\label{alg:classifier_2}
	\begin{algorithmic}[1]
   \STATE {\bfseries Input:} probabilities $\Prob(Y_i = 1 \given \vec{x})= h_i(\vec{x})$, $\forall i \in [m]$, penalty $f(.)$\;
   \STATE {\bfseries Input:} constant $c \geq 0 $\;
   \STATE $s \defi  \left\{ s_i \defi h_i(\vec{x} ) \, | \, i \in  [m]\right\}$\;
   \STATE Sort $s$ in decreasing order: $s_{(1)} \geq s_{(2)} \geq \cdots \geq s_{(m)}$ \;
   \STATE $K_0 \defi \emptyset$, $\mathbf{E}_0 \defi c$ \;
   \STATE $K_2 \defi \lAngle 1, m \rAngle$, $a \defi 1$, $b \defi m$\;
   \STATE $\mathbf{E}_2= \mathbf{E}(\pi_{K_2}) +f(m-2)$ \;
   \FOR{$i=3$ {\bfseries to} $m$}
   \STATE $K_l \defi \lAngle a+1, b \rAngle$ \;
   \STATE $K_r \defi \lAngle a, b-1 \rAngle$  \;
   \IF{$\mathbf{E}(\pi_{K_l}) < \mathbf{E}(\pi_{K_r})$}
   \STATE $K_i \defi K_l$, $a \defi a +1$\;
   \STATE  $\mathbf{E}_i \defi  \mathbf{E}(\pi_{K_l}) +f(m-i)$\;
   \ELSE
   \STATE $K_i \defi K_r$, $b \defi b - 1$\;
   \STATE  $\mathbf{E}_i \defi  \mathbf{E}(\pi_{K_r}) +f(m-i)$ \;
   \ENDIF
   \ENDFOR
   \STATE Determine $d= \operatorname*{argmin}_{i} \mathbf{E}_i$\;
   \STATE {\bfseries Output:} the ranking $\pi_{K_d}$ \;
   \end{algorithmic}
   \end{algorithm}
Lemma \ref{pro:lpp} implies that $K_2 := \lAngle (1), (m) \rAngle$ is an optimal $2$-selection. At each iterative $i = 3, \ldots, m$, Alg. \ref{alg:classifier_2} iteratively looks for the optimal $i$-selection which is either the extensions $\lAngle a+1,b \rAngle$ or $\lAngle a,b-1 \rAngle$ of the optimal $(i-1)$-selection $K_{i-1}:=\lAngle a,b \rAngle$ as claimed in the lemma \ref{pro:lpp2}. The risk-minimizer is simply the optimal $\hat{i}$-selection minimizing the expected loss. 

\section{Proof of Remark \ref{pro:rankcal}}

The proof is carried out with a counter example. Let $m=4$ and $\vec{x}$ be a query instance with the conditional probabilities and the corresponding degrees of uncertainty 
\begin{align*}
\Prob_{\vec{x}} = (0.9, 0.8, 0.7, 0.3)\,,\\
u_{\vec{x}} =  (0.2, 0.4, 0.6, 0.6) \, .
\end{align*}
The extension \eqref{eq:glosscount} of the rank loss is specified by $f:=|A(\hat{\vec{y}})|\cdot c$. The information given by running the algorithm \ref{alg:classifier_2} is presented in Table \ref{tab:infor}.

Let the cost of abstention $c: = 0.03$, thus the risk-minimizing rank is $\{1,4\}$. The risk-minimizer is clearly not uncertainty-aligned since we include the $4$-th label with the degree of uncertainty of $0.6$ while abstain the second label with degree of uncertainty of~$0.4$.

\begin{table}[t]
\caption{Risk-minimizing rank information}
\label{tab:infor}
\vskip 0.15in
\begin{center}
\begin{small}
\begin{sc}
\begin{tabular}{lcccr}
\toprule
$d$ &  $D_d(\hat{\vec{y}}_d)$ & $\mathbf{E}(\pi_{D_d})$ & $f(m-d)$& $\mathbf{E}$ \\
\midrule
$0$ & $\emptyset$   &$0$      &$c\cdot 4$  &$c\cdot 4$\\
$2$ &$\{1,4\}$      &$0.03$   &$c\cdot 2$  &$0.03 + c\cdot 2$ \\
$3$ &$\{1,2,4\}$    &$0.17$   &$c$         &$0.17 + c$ \\
$4$ &$\{1,2,3,4\}$  &$0.47$   &$0$         &$0.47$ \\
\bottomrule
\end{tabular}
\end{sc}
\end{small}
\end{center}
\vskip -0.1in
\end{table}

\section{Proof of Remark \ref{pro:1F}} 
For simplicity, let us write the generalized F-measure \eqref{eq:glF} as
\begin{align}\label{eq:tpp}
F_G(\vec{y}, \hat{\vec{y}}) &=  \frac{2\sum_{i \in D(\hat{\vec{y}})} y_i \hat{y}_i}{\sum_{i \in D(\hat{\vec{y}})} (y_i + \hat{y}_i)} -  f (a) \\
&  = \frac{2 \, t_p}{p + p_p} - f(a)  \, .\nonumber
\end{align}
Turning an incorrect prediction into a correct prediction either means correcting a false positive or a false negative. In the first case, \eqref{eq:tpp} is replaced by $2 t_p/(p+p_p-1) -f(a)$, in the second case by $2(t_p+1)/(p + p_p+1) - f(a)$. In both cases, the value of the measure increases. 

Turning an abstention into a correct prediction either means adding a true positive or a true negative while reducing the abstained size by $1$. In the first case, \eqref{eq:tpp} is replaced by $2 (t_p+2)/(p+p_p+2) -f(a-1)$, in the second case by $2t_p/(p + p_p) - f(a-1)$. In both cases, the value of the measure increases. 

To see that the measure may decrease when turning an incorrect prediction into an abstention, consider the case where $t_p = 0$ and a false negative is turned into an abstention. In this case, \eqref{eq:tpp} is replaced by $2t_p / (p+p_p) - f(a+1)$, which is strictly smaller if $f$ is strictly increasing. 

\section{Proof of Lemma \ref{lem:F2}} 
Consider a prediction $\hat{\vec{y}}$ for a given instance $\vec{x}$, and let $D = D(\hat{\vec{y}})$. Since the expectation of the generalized F-measure is given by 
$$
\mathbf{E} \left(F_G(\vec{y}, \hat{\vec{y}}) \right)  = \mathbf{E} \left(F(\vec{y}_D, \hat{\vec{y}}_D) \right) - f (a) 
$$
and $f (a)$ is a constant, we only need to consider $\mathbf{E} (F(\vec{y}_D, \hat{\vec{y}}_D))$. Exploiting conditional independence of the labels, we can write this expectation as follows:
\begin{align}
\mathbf{E} &\left(F(\vec{y}_D, \hat{\vec{y}}_D) \right) = \sum_{\vec{y}_D \in \mathcal{Y}_D} F(\vec{y}_D, \hat{\vec{y}}_D) \, \Prob(\vec{y}_D \given \vec{x}) \label{eq:htw} \\
&= \sum_{\vec{y}_D \in \mathcal{Y}_D} F(\vec{y}_D, \hat{\vec{y}}_D) \prod_{i \in D} p_i^{y_i} (1- p_i)^{1- y_i} \nonumber  \\
&= \sum_{\vec{y}_D \in \mathcal{Y}_D}  \frac{2\sum_{i \in D} y_i \hat{y}_i}{\sum_{i \in D}( y_i +\hat{y}_i) } \prod_{i \in D} p_i^{y_i} (1- p_i)^{1- y_i} \, . \nonumber
\end{align}
Now, fix some $j \in  D$ and denote the remaining indices on which a prediction is made by $D_j = D \setminus \{ j \}$. Moreover, we use the shorthand notation
\begin{align*}
\alpha(\vec{y}_{D_j}) \defi \prod_{i \in D_j} p_i^{y_i} (1- p_i)^{1- y_i} \, .
\end{align*}
We consider the case where $\hat{y}_j = 1$. 
In the sum \eqref{eq:htw}, which is over all $\vec{y}_D \in \mathcal{Y}_D$, we can separate the cases $\vec{y}_D$ with $y_j = 1$ from those with $y_j = 0$, which yields
\begin{align*}
& \sum_{\vec{y}_{D_j} \in \mathcal{Y}_{D_j}} p_j \alpha(\vec{y}_{D_j}) \left( \frac{2 + 2\sum_{i \in D_j} y_i \hat{y}_i}{2 + \sum_{i \in D_j}( y_i +\hat{y}_i )}  \right) \\
 &+(1-p_j) \alpha(\vec{y}_{D_j}) \left( \frac{2\sum_{i \in D_j} y_i \hat{y}_i}{1 + \sum_{i \in D_j} (y_i + \hat{y}_i )}  \right) \\
& = \sum p_j \alpha(\vec{y}_{D_j}) \beta(\vec{y}_{D_j})  + 
(1-p_j) \alpha(\vec{y}_{D_j}) \beta'(\vec{y}_{D_j}) \, .
\end{align*}
Since $\beta(\vec{y}_{D_j}) > \beta'(\vec{y}_{D_j})$, this expression is monotone increasing in $p_j$. In a similar way, it is shown that the expectation of the generalized F-measure is monotone decreasing in $p_j$ in the case where $\hat{y}_j = 0$.

\section{Proof of Lemma \ref{lem:F3}} 

Let $(\cdot)$ define an order of the labels such that $p_{(i)} > p_{(i+1)}$, $i\in [m-1]$. For a given $k=0, \ldots m$, denote by 
\begin{align}
\mathcal{Y}^k_{\hat{b}} \defi \{\hat{\vec{y}} \in \mathcal{Y}_{pa} \vert \sum_{i\in D(\hat{\vec{y}})} \hat{y}_i = k, |D(\hat{\vec{y}})| = k+b\}\, ,
\end{align}
the optimization problem \eqref{eq:innerG} is decomposed into
\begin{align*}
\hat{\vec{y}}^k_{\hat{b}} & \defi  \arg\max_{\hat{\vec{y}} \in \mathcal{Y}^k_b } \mathbf{E} \left(F_G(\vec{y}, \hat{\vec{y}}) \right) \, , \\
\hat{\vec{y}}^k & \defi  \arg\max_{\hat{\vec{y}} \in \{\hat{\vec{y}}^k_{\hat{b}} \vert b = 0, \ldots,m-k \}} \mathbf{E} \left(F_G(\vec{y}, \hat{\vec{y}}^k_{\hat{b}}) \right) \, . 
\end{align*}
Lemma \ref{lem:F2} together with Lewis's theorem \cite{lewis:1995} imply that, for any number $b = 0, \ldots,m-k$, the optimal partial prediction $\hat{\vec{y}}^k_{\hat{b}}$ has a decision set $D(\hat{\vec{y}}^k_{\hat{b}})$ consisting of $k$ relevant labels whose marginal probabilities are greater than those of the $b$ irrelevant labels. Thus, the $k$-optimal prediction has its predicted part $D(\hat{\vec{y}}^k_{\hat{b}})$ also consists of $k$ relevant labels whose marginal probabilities are greater than those of the $b$ irrelevant labels.

Now, suppose that, contrary to the claim of the lemma, an $k$-optimal prediction $\hat{\vec{y}}^k$ is not of the form (\ref{eq:dselF}), and cannot be improved when replacing any $(i) \in D(\hat{\vec{y}}^k)$ by a $(j)\in A(\hat{\vec{y}}^k)$. Then we necessarily have at least one of the following cases:
\begin{itemize}
\item[-] (i) $\exists (i) \in D(\hat{\vec{y}}^k)$ and $(j) \in A(\hat{\vec{y}}^k)$ s.t. $\hat{y}_{(i)} = 1$ and $p_{(j)} > p_{(i)}$,
\item[-] (ii) $\exists (i) \in D(\hat{\vec{y}}^k)$ and $(j) \in A(\hat{\vec{y}}^k)$ s.t. $\hat{y}_{(i)} = 0$ and $p_{(j)} < p_{(i)}$.
\end{itemize}
The proof is completed by showing that both (i) and (ii) lead to contradiction:
\begin{itemize}
\item[-] (i) Suppose $\hat{y}_{(i)} = 1$ and $p_{(j)} > p_{(i)}$. According to Lemma \ref{lem:F2}, $\mathbf{E} \left(F(\vec{y}_D^k, \hat{\vec{y}}_D^k) \right)$ is an increasing function of $p_{(i)}$ and is increased when replacing $(i)$ by $(j)$, which is a contradiction.
\item[-] (ii) Suppose $\hat{y}_{(i)}= 0$ and $p_{(j)} < p_{(i)}$. According to Lemma \ref{lem:F2}, $\mathbf{E} \left(F(\vec{y}_D^k, \hat{\vec{y}}_D^k) \right)$ is a decreasing function of $p_{(i)}$ and is increased when replacing $(i)$ by $(j)$, which is again a contradiction.  
\end{itemize}  

\section{Proof of Proposition \ref{pro:maximize_F}} 

Given a query instance $\vec{x}$, assume conditional probabilities $p_i =  \Prob(y_i = 1 \given \vec{x}) = h_i(\vec{x})$ are made available by an MLC predictor $\mathbf{h}$. Assuming (conditional) independence of label probabilities in the sense of \eqref{eq:independence}. In the following, we show that a F-maximizer of the generalized F-measure \eqref{eq:glF} is constructed in time $O(m^3)$. 

Let $\hat{\vec{y}}^k_l$ be the partial prediction with with decision set of the form 
\begin{equation*}
D(\hat{\vec{y}}^k) = \lAngle k,l \rAngle \defi \{1, \ldots , k\} \cup \{l , \ldots , m\} \, ,
\end{equation*} 
and $a \defi |\hat{\vec{y}}^k_l|$. Using the shorthand notation 
\begin{align*}
Q(k,k_1) & \defi \Prob\left(\sum_{i=1}^k y_{(i)} = k_1 \vert \vec{x}\right)\, , \\
  S(k, k_1,l) &\defi \sum_{k_2=0}^{m+1 -l}  \frac{ \Prob\left(\sum_{i=l}^m y_{(i)} = k_2 \vert \vec{x}\right)}{k + k_1 + k_2} \,.
\end{align*}
the expected (generalized) F-measure of $\hat{\vec{y}}^k_l$ is
\begin{align*}
F^k_l  & \defi 2 \sum_{\vec{y} \in \mathcal{Y}}\frac{ \Prob(\vec{y}\vert \vec{x}) \sum_{i =1}^k y_{(i)}}{k + \left( \sum_{i=1}^k y_{(i)} + \sum_{i=l}^m y_{(i)}\right)} - f(a) \\
&=  2\sum_{\substack{0 \leq k_1 \leq k \\ 0 \leq k_2 \leq m +1-l}} \frac{ Q(k,k_1) \Prob\left(\sum_{i=l}^m y_{(i)} = k_2 \vert \vec{x}\right) k_1}{k +  k_1 + k_2 } - f(a) \\
&=  2 \sum_{k_1 =0}^k k_1 Q(k,k_1) \sum_{k_2=0}^{m +1 -l}  \frac{ \Prob\left(\sum_{i=l}^m y_{(i)} = k_2 \vert \vec{x}\right)}{k  + k_1 + k_2} - f(a)\\
& = 2 \sum_{k_1 =0}^k  k_1  Q(k,k_1) S(k, k_1,l) - f(a)\,. 
\end{align*}

To compute $Q(k,k_1)$, we employ a list of lists, as discussed by \citet{decubber:2018}, using double indexing, with $k_1 \in \{-1,0 \ldots , k+1\}$. This data structure can also be initialized via dynamic programming:
\begin{align*}
Q(k,k_1)  = & p_{(k)} \Prob\left(\sum_{i=1}^{k-1} y_{(i)} = k_1-1 \vert \vec{x}\right) \\
&+  \left(1-p_{(k)}\right) \Prob\left(\sum_{i=1}^{k-1} y_{(i)} = k_1\vert \vec{x}\right)\\
 = & p_{(k)} Q(k-1,k_1-1) + \left(1-p_{(k)}\right) Q(k-1,k_1) \nonumber \,,
\end{align*}
with the boundary conditions 
\begin{align*}
& Q(1,:)  = \left(0, 1-p_{(1)},p_{(1)},0\right) \, ,\\
& Q(k,-1) = L(k,k+1) =0 \, .
\end{align*}

For any fixed number $k \in [m]$, $S(k, k_1,l)$ can be computed via the following recursive relation:
\begin{align*}
S(k, k_1,l) = & p_{(l)}\sum_{k_2=1}^{m +1 -l}  \frac{ \Prob\left(\sum_{i=l+1}^m y_{(i)} = k_2 -1 \vert \vec{x}\right)}{k  + k_1+ k_2}\\
& + \left(1-  p_{(l)}\right)\sum_{k_2=0}^{m -l}  \frac{ \Prob\left(\sum_{i=l+1}^m y_{(i)} = k_2\vert \vec{x}\right)}{k  + k_1 + k_2} \\
 = & p_{(l)}\sum_{k_2=0}^{m -l}  \frac{ \Prob\left(\sum_{i=l+1}^m y_{(i)} = k_2\vert \vec{x}\right)}{k  + (k_1+1) + k_2} \\
 &+ \left(1-  p_{(l)}\right)\sum_{k_2=0}^{m -l}  \frac{ \Prob\left(\sum_{i=l+1}^m y_{(i)} = k_2\vert \vec{x}\right)}{k  + k_1 + k_2} \\
 =& p_{(l)}\sum_{k_2=0}^{m +1 -(l+1)}  \frac{ \Prob\left(\sum_{i=l+1}^m y_{(i)} = k_2\vert \vec{x}\right)}{k   + (k_1+1) + k_2} \\
 &+ \left(1-  p_{(l)}\right)\sum_{k_2=0}^{m+1 -(l+1)}  \frac{ \Prob\left(\sum_{i=l+1}^m y_{(i)} = k_2\vert \vec{x}\right)}{k  + k_1 + k_2} \\
= & p_{(l)} S(k, k_1+1,l+1) \\
&+ \left(1-  p_{(l)}\right)S(k, k_1,l+1)\,,
\end{align*}
with the boundary conditions
\begin{align*}
S(k, k_1,m+1) = \frac{1}{k  + k_1} \, , \forall (k, k_1) \, ,
\end{align*} 
where $l = m+1$ means there is no irrelevant label. Altogether, we come up to the implementation given in Algorithm \ref{alg:GFMaximizer} which requires the time $O(m)^3$. 

\begin{algorithm}[t]
	\caption{Determining a F-maximizer of the generalized F-measure}\label{alg:GFMaximizer}
\begin{algorithmic}[1]
   \STATE {\bfseries Input:} marginal probabilities $\vec{\Prob} =(p_1, p_2, \ldots, p_m)$, penalty $f(\cdot)$
   \STATE $\vec{\Prob}\longleftarrow \textbf{sort}(\vec{\Prob})$ s.t. $p_1 \geq p_2 \geq \ldots, \geq p_m$
   \STATE \textbf{initialize} $Q$ $\longleftarrow$ a list of $m$ empty lists 
   \STATE \textbf{set} $Q(1,:) = \left(0,(1-p_1), p_1,0\right)$ \\
   \FOR{$k=2$ {\bfseries to} $m$}
       \STATE $Q(k,:)$ $\longleftarrow$ a list of $k+3$ zeros with index starting at $-1$
       \FOR{$j=0$ {\bfseries to} $k$}
           \STATE $Q(k,j)\longleftarrow p_k Q(k-1,j-1) + (1-p_k) Q(k-1,j)$
       \ENDFOR
   \ENDFOR
   \STATE  $F^0_{m+1} \longleftarrow  1 - f(m)$
   \STATE  $l_0 \longleftarrow  m+1$ 
   \FOR{$k=1$ {\bfseries to} $m$}
       \FOR{$i=0$ {\bfseries to} $m$}
           \STATE \textbf{initialize} $S(k, i) \longleftarrow  \frac{1}{k  + i}$
       \ENDFOR
       \STATE  $F^k_{m+1} \longleftarrow 2 \sum_{k_1 =0}^k k_1  Q(k,k_1) S(k,k_1) - f(m-k)$
       \FOR{$l=m$ {\bfseries to} $k+1$}
           \FOR{$i=0$ {\bfseries to} $l-1$}
               \STATE \textbf{update} $S(k, i) \longleftarrow  p_{(l)} S(k, i+1) + \left(1-  p_{(l)}\right)S(k, i)$
           \ENDFOR
           \STATE  $F^k_l \longleftarrow 2  \sum_{k_1 =0}^k k_1  Q(k,k_1) S(k,k_1) - f(l-k-1)$ 
       \ENDFOR
       \STATE  $l_k \longleftarrow  \operatorname*{argmax}_{l} F^k_l$ , $F^k \longleftarrow  F^k_{l_k}$ 
   \ENDFOR
   \STATE  $k^* \longleftarrow \operatorname*{argmax}_{k} F^k$
   \STATE {\bfseries Output:}  F-maximizer $\hat{\vec{y}} \defi \hat{\vec{y}}^{k^*}_{l_{k^*}}$ 
\end{algorithmic}
\end{algorithm}

\section{Additional Experiments}

\subsection{The Cases of Hamming Loss and Rank Loss}

\begin{figure}
\centering 
\SetFigLayout{1}{1}{
\includegraphics{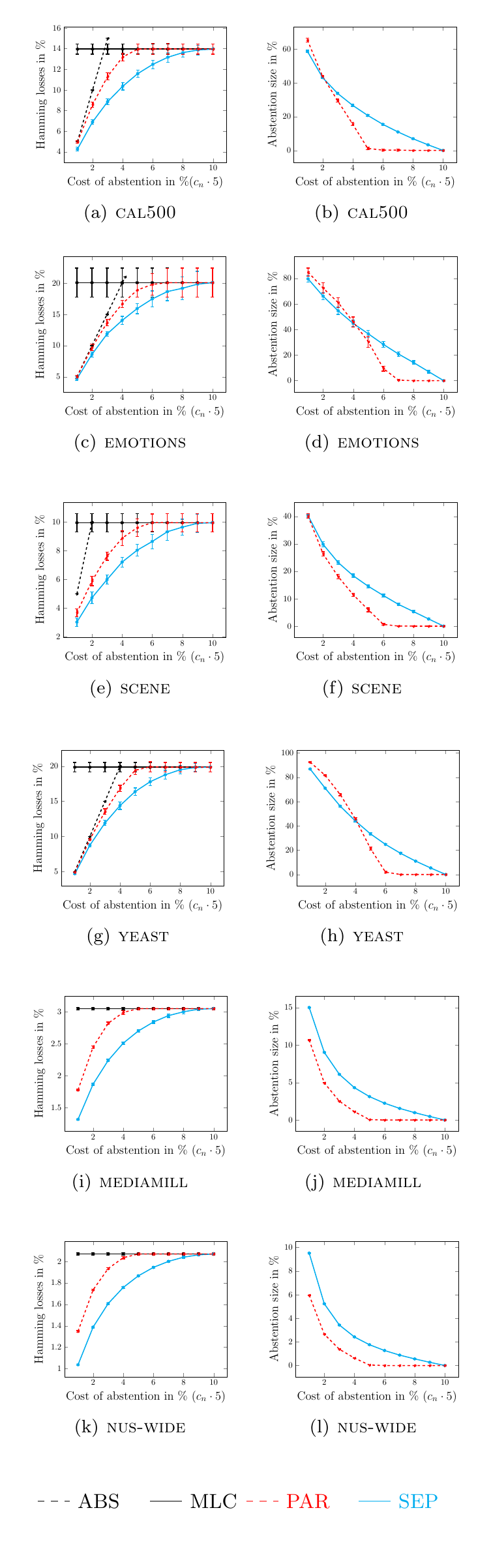}}
\caption{Binary relevance with logistic regression: Experimental results in terms of expected Hamming loss $(L_H \cdot 100)/m$ and abstention size (in percent) for $f_1(a) = a \cdot c$ ({\sc SEP}) and $f_2(a) = (a\cdot m\cdot c)/(m+a)$ ({\sc PAR}), as a function of the cost of abstention.}
\label{fig:br_lr_h}  
\end{figure}

\begin{figure}
\centering 
\SetFigLayout{1}{1}{
\includegraphics{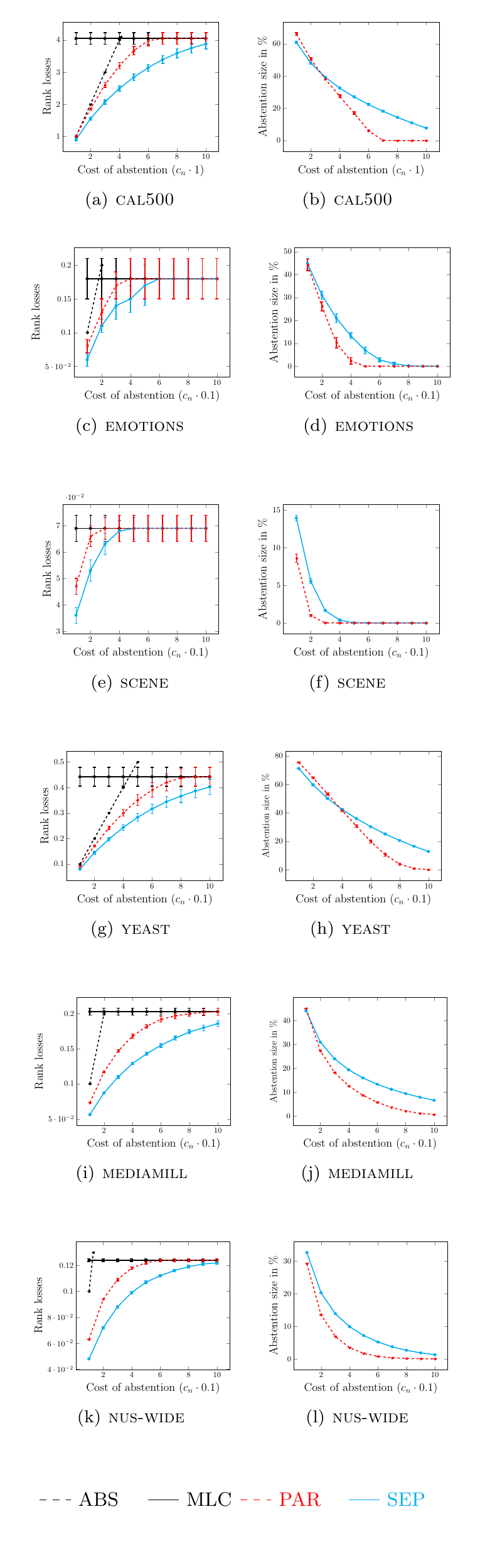}}
\caption{Binary relevance with logistic regression: Experimental results in terms of expected rank loss $L_R/m$ and abstention size (in percent) for $f_1(a) = a \cdot c$ ({\sc SEP}) and $f_2(a) = (a\cdot m\cdot c)/(m+a)$ ({\sc PAR}), as a function of the cost of abstention.}
\label{fig:br_lr_r}  
\end{figure}

In addition to the experiments which presented in Section \ref{sec:exp} (illustrated in Figures \ref{fig:br_lr_h} and \ref{fig:br_lr_r} for all the six data sets), we conduct three series of experiments with other based learners. Similar to what has been done in Section \ref{sec:exp}, we compare the performance of reliable classifiers (SEPH and PARH) to the conventional classifier that makes full predictions ({\sc BRH}) as well as the cost of full abstention ({\sc ABSH}).

\noindent \textit{Classifier Chains Learning with Logistic Regression}

We use classifier chains (CC) learning \citep{read:2009} with logistic regression (LR) as base learner, where we train $m$ LR classifiers ordered in a chain according to the Bayesian chain rule. The first classifier is trained just on the input space, and then each next classifier is trained on the input space and all previous classifiers in the chain.

\begin{figure}
\centering 
\SetFigLayout{1}{1}{
\includegraphics{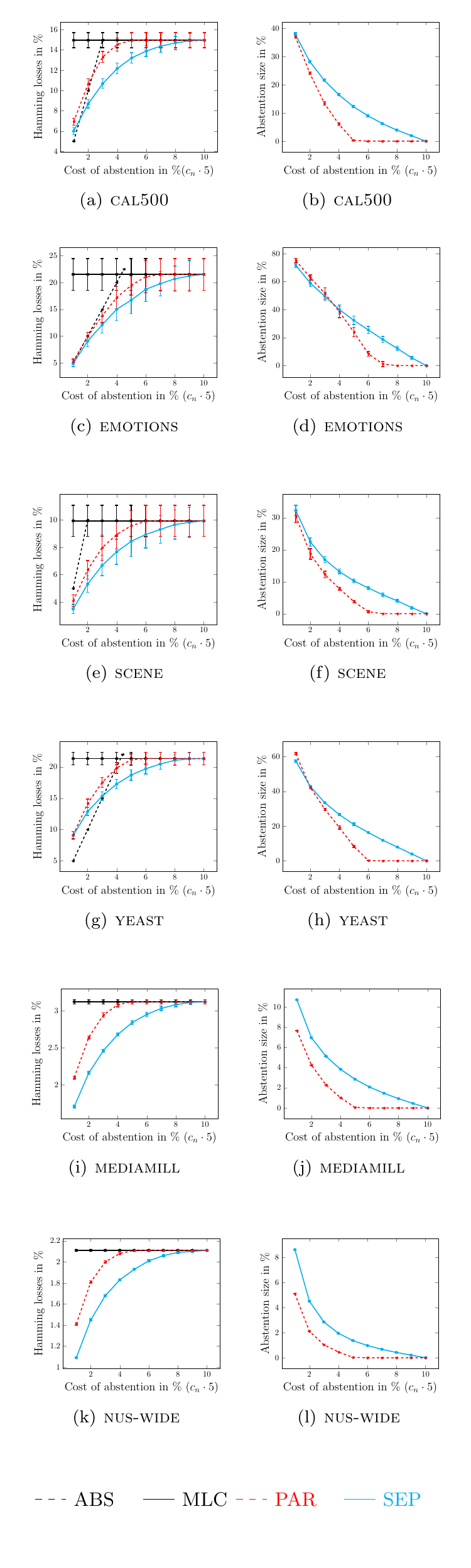}}
\caption{Classifier chains with logistic regression: Experimental results in terms of expected Hamming loss $(L_H \cdot 100)/m$ and abstention size (in percent) for $f_1(a) = a \cdot c$ ({\sc SEP}) and $f_2(a) = (a\cdot m\cdot c)/(m+a)$ ({\sc PAR}), as a function of the cost of abstention.}
\label{fig:cc_lr_h}  
\end{figure}

\begin{figure}
\centering 
\SetFigLayout{1}{1}{
\includegraphics{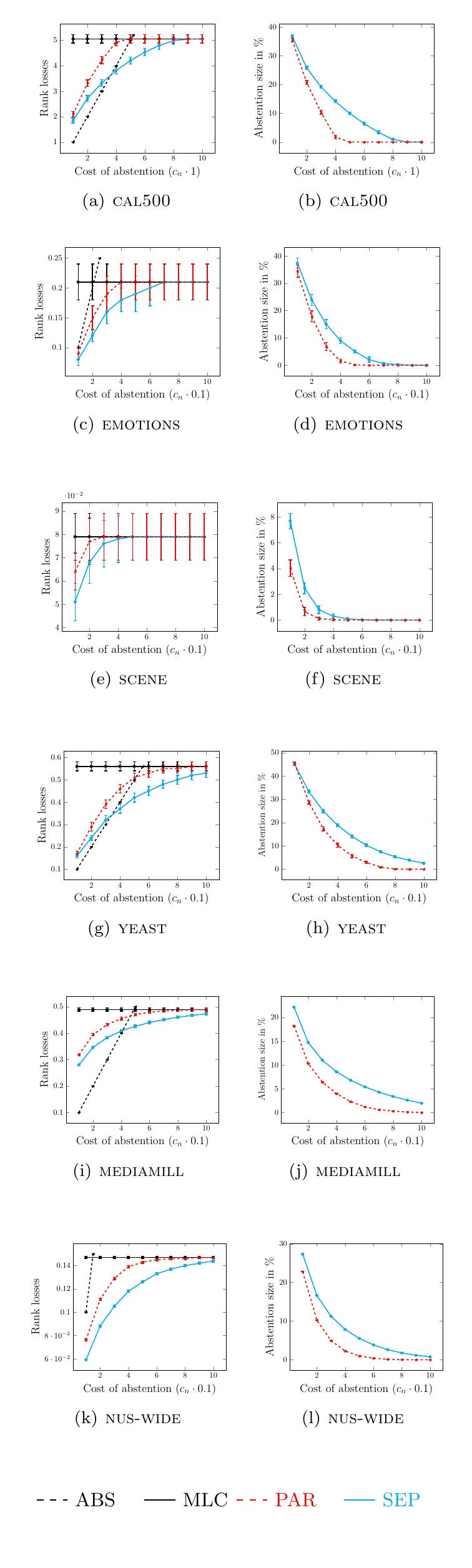}}
\caption{Classifier chains with logistic regression: Experimental results in terms of expected rank loss $L_R/m$ and abstention size (in percent) for $f_1(a) = a \cdot c$ ({\sc SEP}) and $f_2(a) = (a\cdot m\cdot c)/(m+a)$ ({\sc PAR}), as a function of the cost of abstention.}
\label{fig:cc_lr_r}  
\end{figure}

We used the default in skmultilearn, i.e., with regularisation parameter of LR set to $1$ and the classifier chains follow the same ordering as provided in the training set, i.e. label in column $0$, then $1$, and so on \footnote{For an implementation in Python, see \url{http://scikit.ml/api/skmultilearn.problem_transform.cc.html}.}. The results are illustrated in Figures \ref{fig:cc_lr_h} and \ref{fig:cc_lr_r}.

\noindent \textit{Binary Relevance and Classifier Chains Learning with SVM}

\begin{figure}
\centering 
\SetFigLayout{1}{1}{
\includegraphics{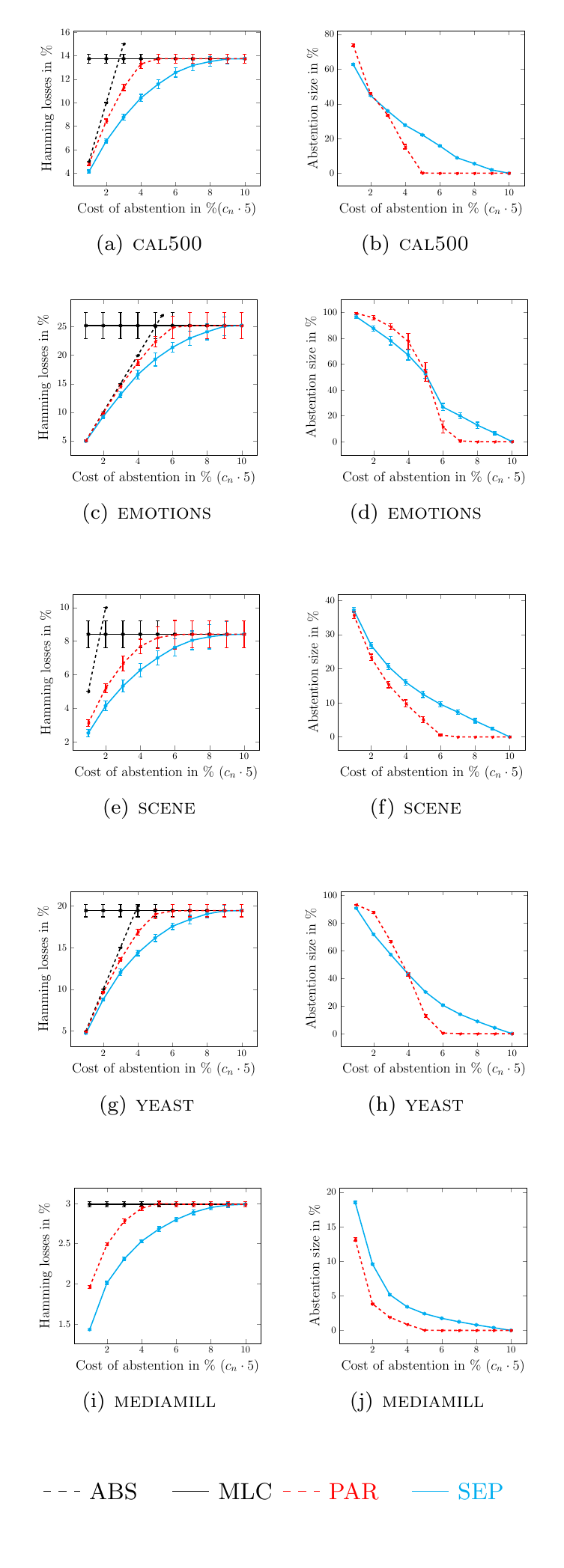}}
\caption{Binary relevance with SVM: Experimental results in terms of expected Hamming loss $(L_H \cdot 100)/m$ and abstention size (in percent) for $f_1(a) = a \cdot c$ ({\sc SEP}) and $f_2(a) = (a\cdot m\cdot c)/(m+a)$ ({\sc PAR}), as a function of the cost of abstention.}
\label{fig:br_svm_h}  
\end{figure}

\begin{figure}
\centering 
\SetFigLayout{1}{1}{
\includegraphics{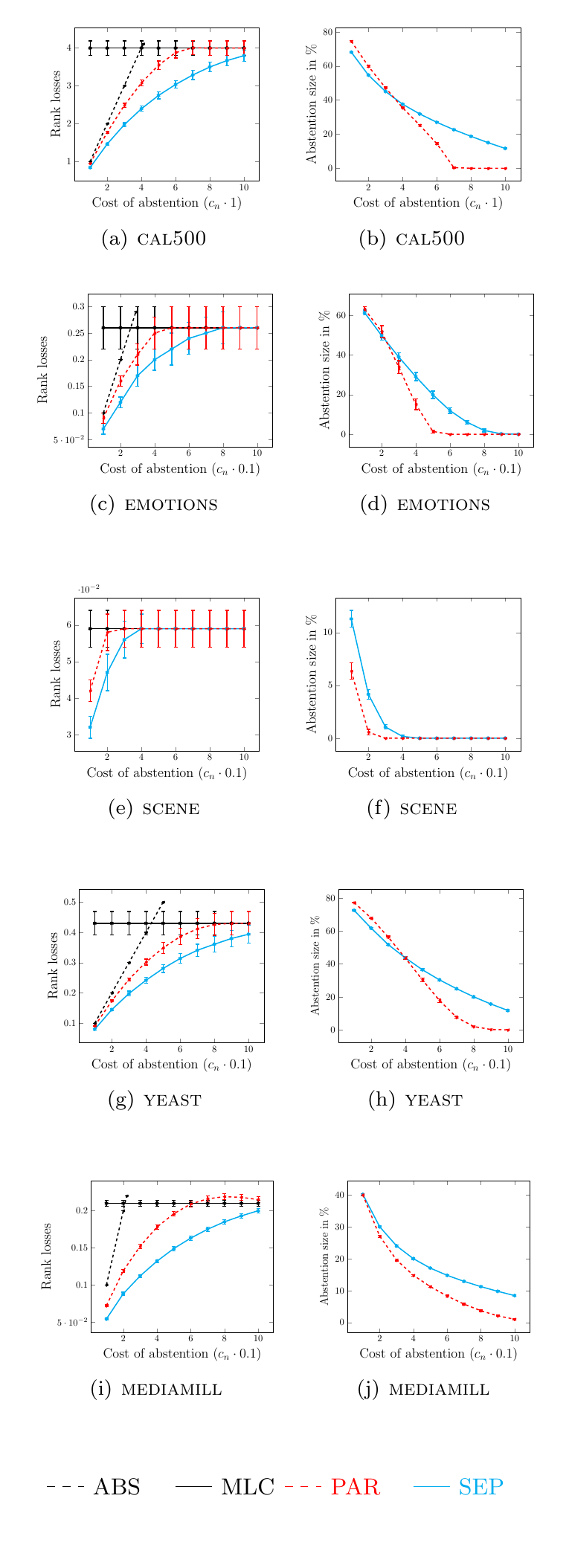}}
\caption{Binary relevance with SVM: Experimental results in terms of expected rank loss $L_R/m$ and abstention size (in percent) for $f_1(a) = a \cdot c$ ({\sc SEP}) and $f_2(a) = (a\cdot m\cdot c)/(m+a)$ ({\sc PAR}), as a function of the cost of abstention.}
\label{fig:br_svm_r}  
\end{figure}

\begin{figure}
\centering 
\SetFigLayout{1}{1}{
\includegraphics{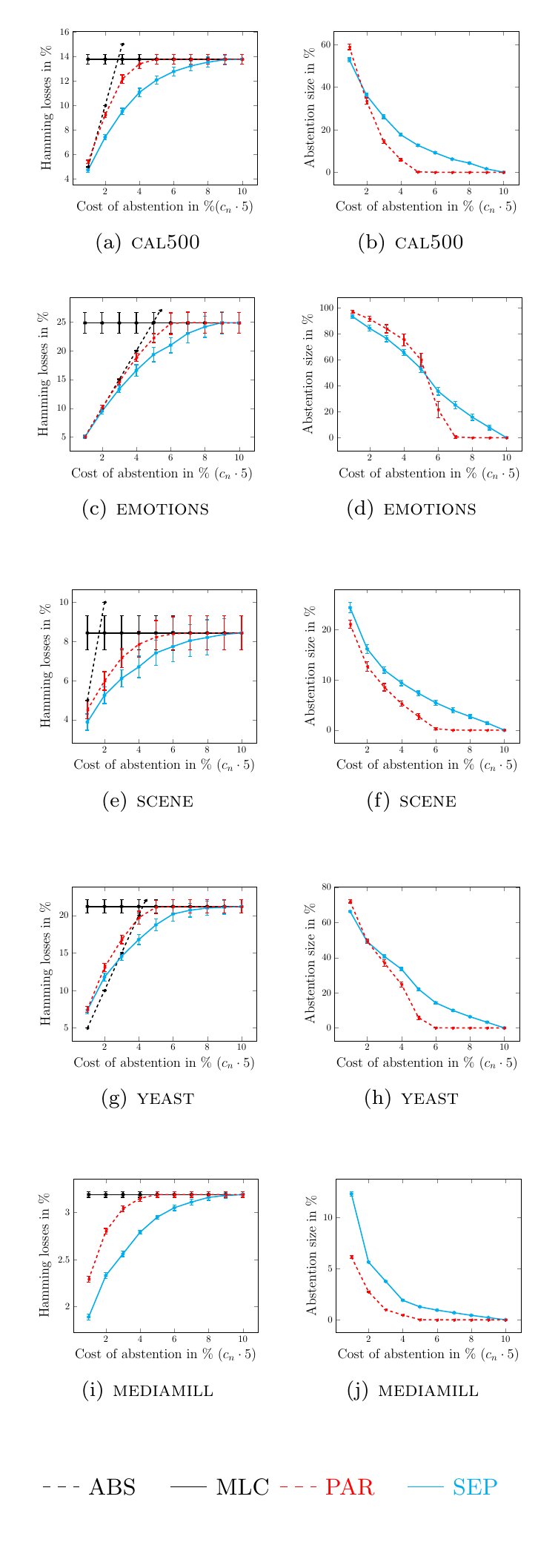}}
\caption{Classifier chains with SVM: Experimental results in terms of expected Hamming loss $(L_H \cdot 100)/m$ and abstention size (in percent) for $f_1(a) = a \cdot c$ ({\sc SEP}) and $f_2(a) = (a\cdot m\cdot c)/(m+a)$ ({\sc PAR}), as a function of the cost of abstention.}
\label{fig:cc_svm_h}  
\end{figure}

\begin{figure}
\centering 
\SetFigLayout{1}{1}{
\includegraphics{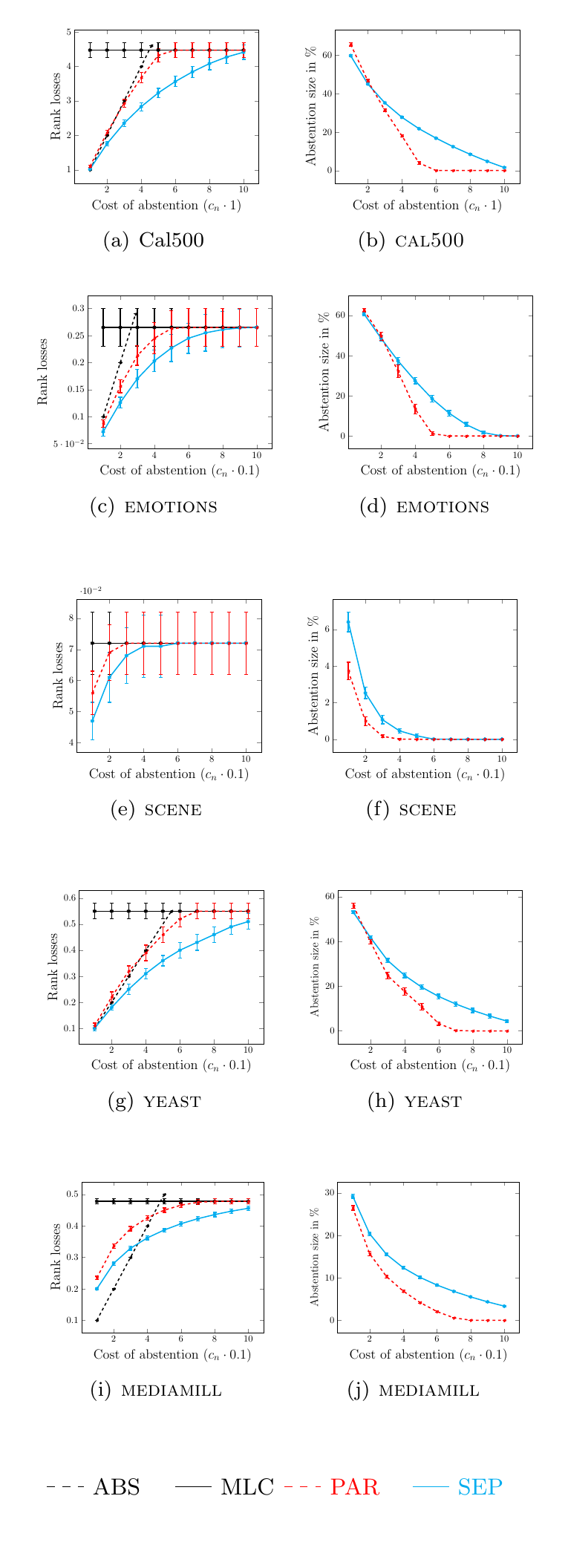}}
\caption{Classifier chains with SVM: Experimental results in terms of expected rank loss $L_R/m$ and abstention size (in percent) for $f_1(a) = a \cdot c$ ({\sc SEP}) and $f_2(a) = (a\cdot m\cdot c)/(m+a)$ ({\sc PAR}), as a function of the cost of abstention.}
\label{fig:cc_svm_r}  
\end{figure}

We also conduct two series of experiments following the BR and CC learning with SVM as base learner. Note that the standard SVMs do not provide probabilistic predictions, we train the parameters of an additional sigmoid function to map the SVM outputs into probabilities \citep{lin:2007,platt:1999} using an internal five-fold cross validation. The results for the BR and CC learning are illustrated in the figure \ref{fig:br_svm_h}--\ref{fig:br_svm_r}, and \ref{fig:cc_svm_h}--\ref{fig:cc_svm_r}, respectively.

\subsection{The Case of F-measure}

\begin{figure}
\centering 
\includegraphics[width=\linewidth]{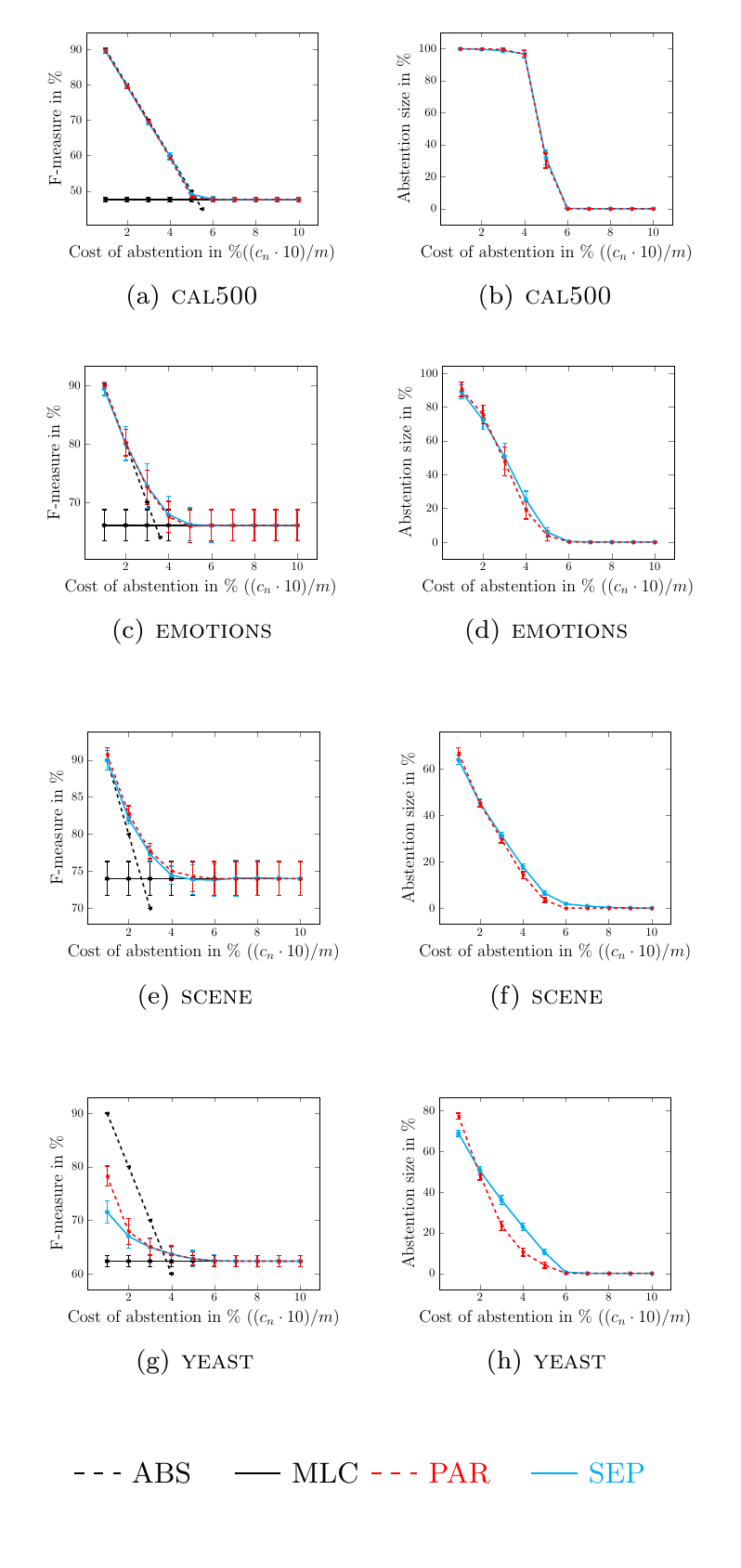}
\caption{Binary relevance with logistic regression: Experimental results in terms of expected F-measure and abstention size (in percent) for $f_1(a) = a \cdot c$ ({\sc SEP}) and $f_2(a) = (a\cdot m\cdot c)/(m+a)$ ({\sc PAR}), as a function of the cost of abstention.}
\label{fig:f_br_lr}  
\end{figure}

\begin{figure}
\centering 
\includegraphics[width=\linewidth]{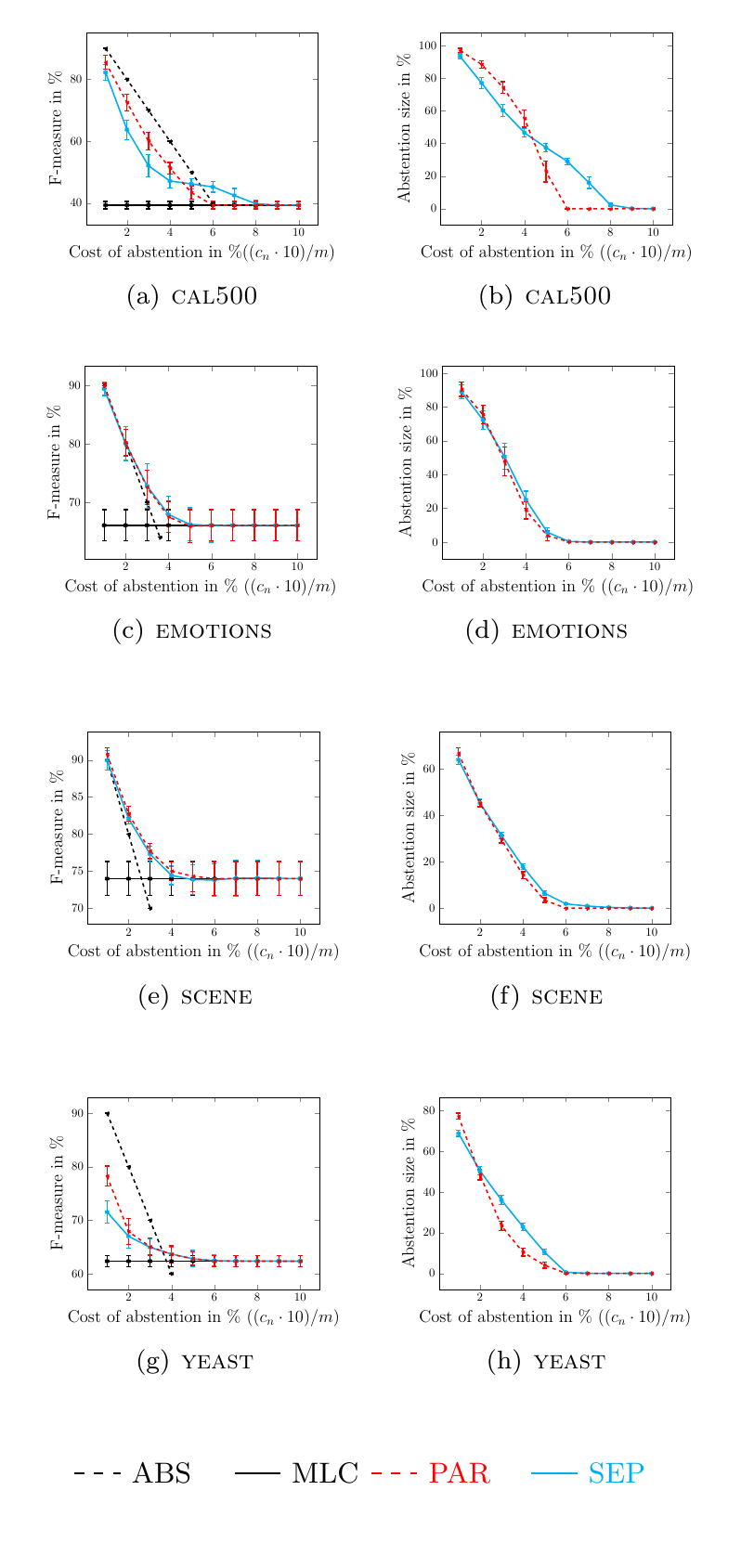}
\caption{Classifier chains with logistic regression: Experimental results in terms of expected F-measure and abstention size (in percent) for $f_1(a) = a \cdot c$ ({\sc SEP}) and $f_2(a) = (a\cdot m\cdot c)/(m+a)$ ({\sc PAR}), as a function of the cost of abstention.}
\label{fig:f_cc_lr}  
\end{figure}

We conduct two series of experiments following the BR and CC learning with logistic regression as base learner. The results for the BR and CC learning are illustrated in the figure \ref{fig:f_br_lr} and \ref{fig:f_cc_lr}, respectively.

\newpage
\clearpage 

\end{appendix}

\end{document}